\documentclass[10pt,twocolumn,letterpaper]{article}

\usepackage{cvpr}
\usepackage{times}
\usepackage{epsfig}
\usepackage{graphicx}
\usepackage{amsmath}
\usepackage{amssymb}
\usepackage{caption}
\usepackage{subcaption}
\usepackage{latexsym}
\usepackage{amsfonts}
\usepackage{color}
\usepackage{soul}
\usepackage{appendix}
\usepackage{url}

\usepackage[pagebackref=true,breaklinks=true,letterpaper=true,colorlinks,bookmarks=false]{hyperref}

\newcommand{\cev}[1]{\reflectbox{\ensuremath{\vec{\reflectbox{\ensuremath{#1}}}}}}

\cvprfinalcopy 

\def\cvprPaperID{****} 

\begin{document}

\title{MAttNet: Modular Attention Network for Referring Expression Comprehension}

\author{Licheng Yu$^1$, Zhe Lin$^2$, Xiaohui Shen$^2$, Jimei Yang$^2$, Xin Lu$^2$,\\
Mohit Bansal$^1$, Tamara L. Berg$^1$\\ \\
$^1$University of North Carolina at Chapel Hill
~~~$^2$Adobe Research
}

\maketitle

\begin{abstract}
In this paper, we address referring expression comprehension: localizing an image region described by a natural language expression.
While most recent work treats expressions as a single unit, 
we propose to decompose them into three modular components related to subject appearance, location, and relationship to other objects. This allows us to flexibly adapt to expressions containing different types of information in an end-to-end framework.
In our model, which we call the Modular Attention Network (MAttNet), two types of attention are utilized: language-based attention that learns the module weights as well as the word/phrase attention that each module should focus on; and visual attention that allows the subject and relationship modules to focus on relevant image components. 
Module weights combine scores from all three modules dynamically to output an overall score.
Experiments show that MAttNet outperforms previous state-of-the-art methods by a large margin on both bounding-box-level and pixel-level comprehension tasks.
Demo\footnote{Demo: \href{http://vision2.cs.unc.edu/refer/comprehension}{vision2.cs.unc.edu/refer/comprehension}} and code\footnote{Code: \href{https://github.com/lichengunc/MAttNet}{https://github.com/lichengunc/MAttNet}} are provided.
\end{abstract}


\section{Introduction}
\label{sec:introduction}
\vspace{-.2cm}

\begin{figure}[t]
\centering
\includegraphics[width=0.5\textwidth]{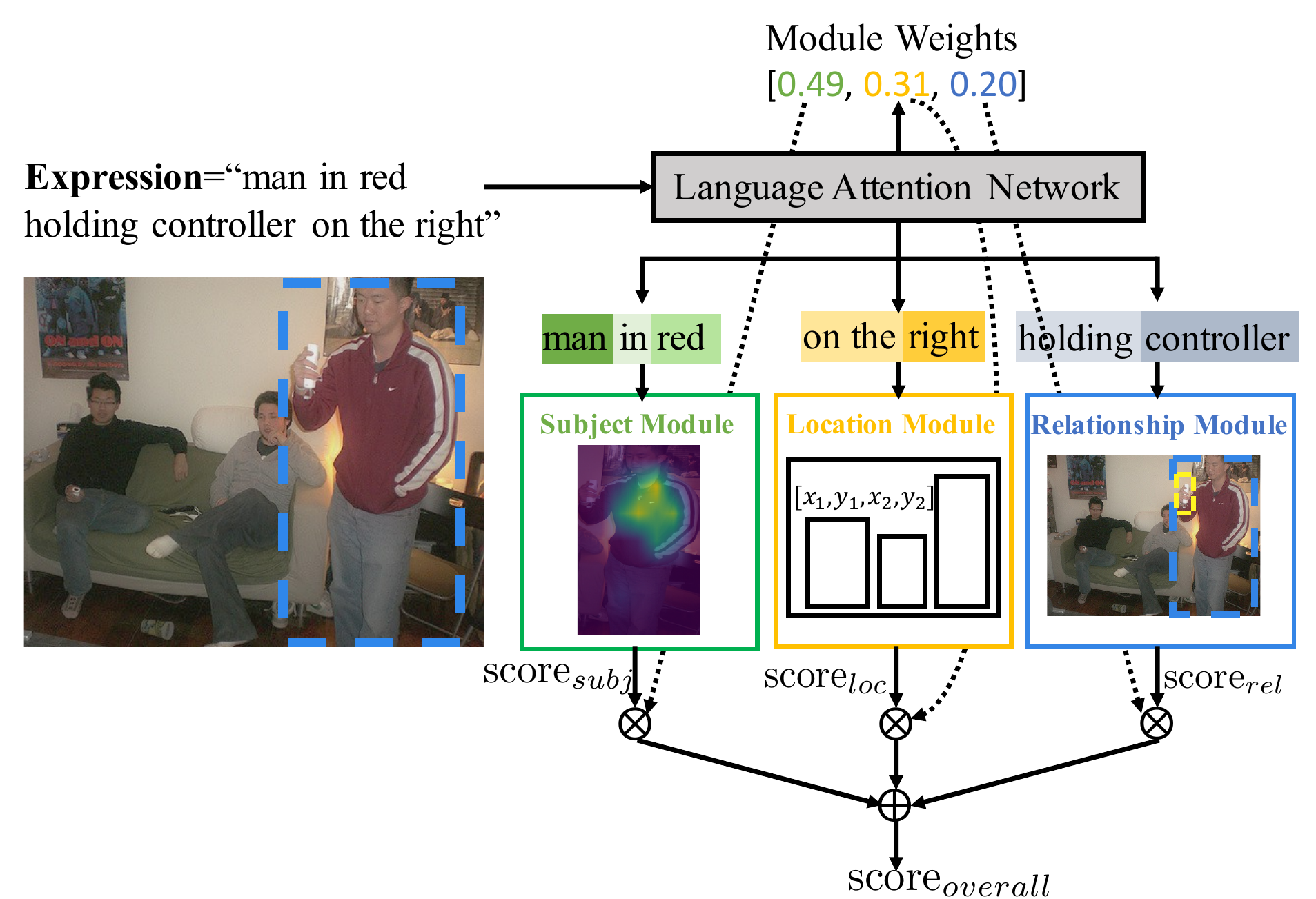}
\caption{Modular Attention Network (MAttNet). Given an expression, we attentionally parse it into three phrase embeddings, which are input to three visual modules that process the described visual region in different ways and compute individual matching scores. An overall score is then computed as a weighted combination of the module scores.
}
\label{fig:model}
\end{figure}

Referring expressions are natural language utterances that indicate particular objects within a scene, e.g., ``the woman in the red sweater'' or ``the man on the right''. For robots or other intelligent agents communicating with people in the world, the ability to accurately comprehend such expressions in real-world scenarios will be a necessary component for natural interactions.  

Referring expression comprehension is typically formulated as selecting the best region from a set of proposals/objects $O=\{o_i\}_{i=1}^N$ in image $I$, given an input expression $r$. 
Most recent work on referring expressions uses CNN-LSTM based frameworks to model $P(r|o)$~\cite{mao2015generation,hu2015natural,yu2016refer,nagaraja16refexp,luo2017comprehension} or uses a joint vision-language embedding framework to model $P(r, o)$ ~\cite{rohrbach2016grounding,wang2015learning,wang2017learning}.
During testing, the proposal/object with highest likelihood/probability is selected as the predicted region.
However, most of these work uses a simple concatenation of all features (target object feature, location feature and context feature) as input and a single LSTM to encode/decode the whole expression, ignoring the variance among different types of referring expressions. Depending on what is distinctive about a target object, different kinds of information might be mentioned in its referring expression. For example, if the target object is a red ball among 10 black balls then the referring expression may simply say ``the red ball''. If that same red ball is placed among 3 other red balls then location-based information may become more important, e.g., ``red ball on the right''. Or, if there were 100 red balls in the scene then the ball's relationship to other objects might be the most distinguishing information, e.g., ``red ball next to the cat''. Therefore, it is natural and intuitive to think about the comprehension model as a modular network, where different visual processing modules are triggered based on what information is present in the referring expression.

Modular networks have been successfully applied to address other tasks such as (visual) question answering~\cite{andreas2016learning, andreas2016neural}, visual reasoning~\cite{hu2017learning, johnson2017inferring}, relationship modeling~\cite{ronghang16relationship}, and multi-task reinforcement learning~\cite{andreas2016modular}.
To the best our knowledge, we present the first modular network for the general referring expression comprehension task. Moreover, these previous work typically relies on an off-the-shelf language parser~\cite{socher2013parsing} to parse the query sentence/question into different components and dynamically assembles modules into a model addressing the task.
However, the external parser could raise parsing errors and propagate them into model setup, adversely effecting performance.

Therefore, in this paper we propose a modular network for referring expression comprehension - Modular Attention Network (MAttNet) - that takes a natural language expression as input and softly decomposes it into three phrase embeddings. These embeddings are used to trigger three separate visual modules (for subject, location, and relationship comprehension, each with a different attention model) to compute matching scores, which are finally combined into an overall region score based on the module weights. Our model is illustrated in Fig.~\ref{fig:model}. There are 3 main novelties in MAttNet.

First, MAttNet is designed for general referring expressions.
It consists of 3 modules: subject, location and relationship.
As in~\cite{kazemzadeh2014referitgame}, a referring expression could be parsed into 7 attributes: category name, color, size, absolute location, relative location, relative object and generic attribute.
MAttNet covers all of them.
The subject module handles the category name, color and other attributes, the location module handles both absolute and (some) relative location, and the relationship module handles subject-object relations.
Each module has a different structure and learns the parameters within its own modular space, without affecting the others. 

Second, MAttNet learns to parse expressions automatically through a soft attention based mechanism, instead of relying on an external language parser~\cite{socher2013parsing,kazemzadeh2014referitgame}.
We show that our learned ``parser'' attends to the relevant words for each module and outperforms an off-the-shelf parser by a large margin.
Additionally, our model computes module weights which are adaptive to the input expression, measuring how much each module should contribute to the overall score.
Expressions like ``red cat'' will have larger subject module weights and smaller location and relationship module weights, while expressions like ``woman on left'' will have larger subject and location module weights.

Third, we apply different visual attention techniques in the subject and relationship modules to allow relevant attention on the described image portions. 
In the subject module, soft attention attends to the parts of the object itself mentioned by an expression like ``man in red shirt'' or ``man with yellow hat''. We call this ``in-box'' attention. In contrast, in the relationship module, hard attention is used to attend to the relational objects mentioned by expressions like ``cat on chair'' or ``girl holding frisbee''. 
Here the attention focuses on ``chair'' and ``frisbee'' to pinpoint the target object ``cat'' and ``girl''. We call this ``out-of-box'' attention.
We demonstrate both attentions play important roles in improving comprehension accuracy.

During training, the only supervision is object proposal, referring expression pairs, $(o_i, r_i)$, and all of the above are automatically learned in an end-to-end unsupervised manner, including the word attention, module weights, soft spatial attention, and hard relative object attention.

We demonstrate MAttNet has significantly superior comprehension performance over all state-of-the-art methods, achieving ${\sim}10\%$ improvements on bounding-box localization and almost doubling precision on pixel segmentation.

\section{Related Work}
\label{sec:related}
\vspace{-.2cm}

\smallskip
\noindent{\bf Referring Expression Comprehension:}
The task of referring expression comprehension is to localize a region described by a given referring expression.
To address this problem, some recent work\cite{mao2015generation, yu2016refer, nagaraja16refexp, hu2015natural,luo2017comprehension} uses CNN-LSTM structure to model $P(r|o)$ and looks for the object $o$ maximizing the probability.
Other recent work uses joint embedding model~\cite{rohrbach2016grounding, wang2015learning, liu2017referring, chen2017query} to compute $P(o|r)$ directly.
In a hybrid of both types of approaches, ~\cite{yu2016joint} proposed a joint speaker-listener-reinforcer model that combined CNN-LSTM (speaker) with embedding model (listener) to achieve state-of-the-art results.

Most of the above treat comprehension as bounding box localization, but object segmentation from referring expression has also been studied in some recent work~\cite{hu2016segmentation, liu2017recurrent}.
These papers use FCN-style~\cite{long2015fully} approaches to perform expression-driven foreground/background classification. We demonstrate that in addition to bounding box prediction, we also outperform previous segmentation results.

\smallskip
\noindent{\bf Modular Networks:}
Neural module networks~\cite{andreas2016neural} were introduced for visual question answering. These networks decompose the question into several components and dynamically assemble a network to compute an answer to the given question.
Since their introduction, modular networks have been applied to several other tasks: visual reasoning~\cite{hu2017learning, johnson2017inferring}, question answering~\cite{andreas2016learning}, relationship modeling~\cite{ronghang16relationship}, multitask reinforcement learning~\cite{andreas2016modular}, etc.
While the early work~\cite{andreas2016neural, johnson2017inferring, andreas2016learning} requires an external language parser to do the decomposition, recent methods~\cite{ronghang16relationship, hu2017learning} propose to learn the decomposition end-to-end. We apply this idea to referring expression comprehension, also taking an end-to-end approach bypassing the use of an external parser. We find that our soft attention approach achieves better performance over the hard decisions predicted by a parser.

The most related work to us is~\cite{ronghang16relationship}, which decomposes the expression into (Subject, Preposition/Verb, Object) triples.
However, referring expressions have much richer forms than this fixed template.
For example, expressions like ``left dog'' and ``man in red'' are hard to model using~\cite{ronghang16relationship}.
In this paper, we propose a generic modular network addressing all kinds of referring expressions.
Our network is adaptive to the input expression by assigning both word-level attention and module-level weights.

\vspace{-.1cm}
\section{Model}
\label{sec:model}
\vspace{-.1cm}

MAttNet is composed of a language attention network plus visual subject, location, and relationship modules.
Given a candidate object $o_i$ and referring expression $r$, we first use the language attention network to compute a soft parse of the referring expression into three components (one for each visual module) and map each to a phrase embedding.
Second, we use the three visual modules (with unique attention mechanisms) to compute matching scores for $o_i$ to their respective embeddings.
Finally, we take a weighted combination of these scores to get an overall matching score, measuring the compatibility between $o_i$ and $r$.

\vspace{-.1cm}
\subsection{Language Attention Network}\label{sec:lang_model}
\vspace{-.1cm}

\begin{figure}[t]
\centering
\includegraphics[width=0.45\textwidth]{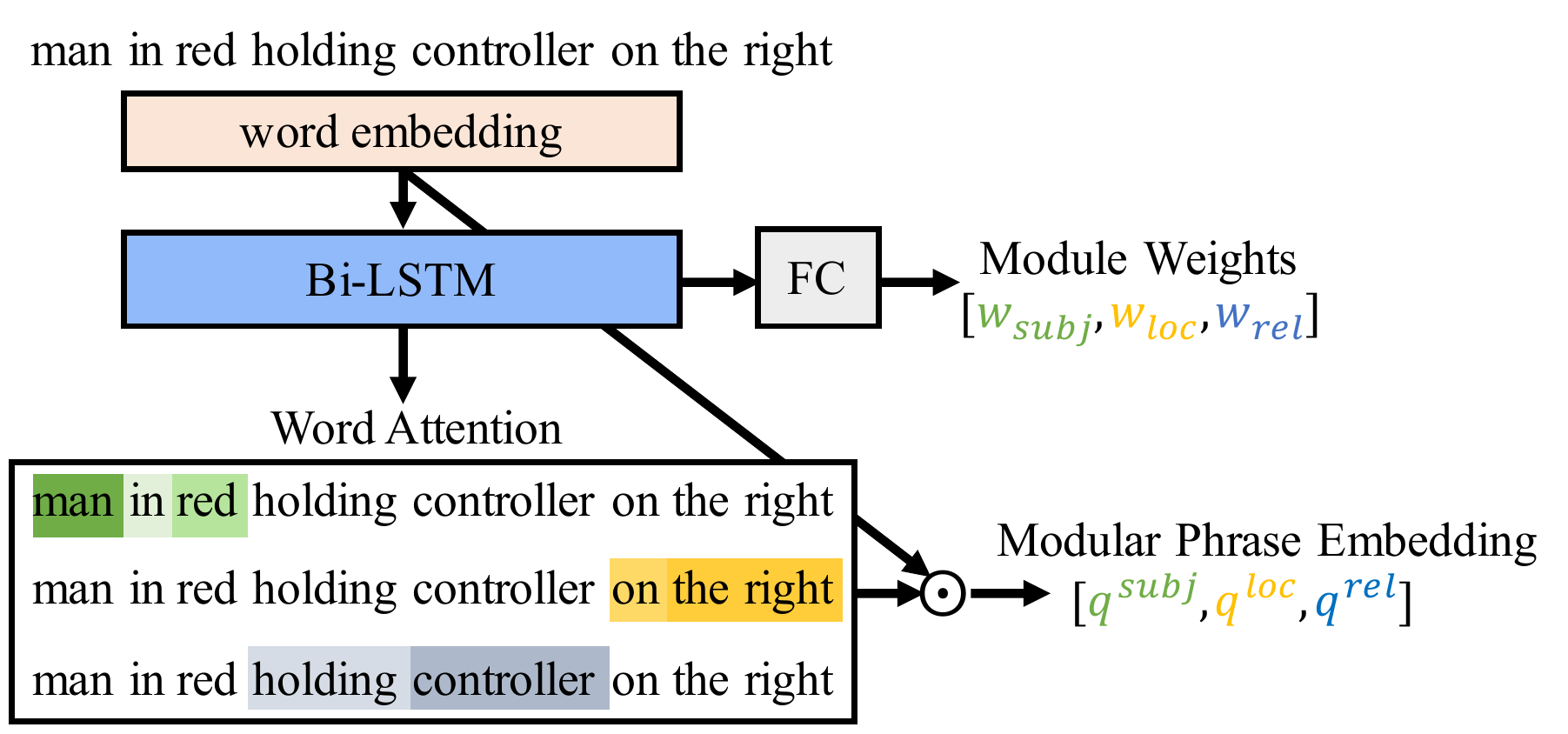}
\vspace{-.1cm}
\caption{Language Attention Network}
\label{fig:lang_net}
\end{figure}

Instead of using an external language parser~\cite{socher2013parsing}\cite{andreas2016neural}\cite{andreas2016learning} or pre-defined templates~\cite{kazemzadeh2014referitgame} to parse the expression, we propose to learn to attend to the relevant words automatically for each module, similar to~\cite{ronghang16relationship}.
Our language attention network is shown in Fig.~\ref{fig:lang_net}.
For a given expression $r=\{u_t\}_{t=1}^T$, we use a bi-directional LSTM to encode the context for each word.
We first embed each word $u_t$ into a vector $e_t$ using an one-hot word embedding, then a bi-directional LSTM-RNN is applied to encode the whole expression.
The final hidden representation for each word is the concatenation of the hidden vectors in both directions:
\vspace{-.1cm}
\begin{equation}
\begin{split}\nonumber
e_t &= \mbox{embedding}(u_t) \\ 
\vec{h}_t &= \vec{\mbox{LSTM}}(e_t, \vec{h}_{t-1}) \\
\cev{h}_t &= \cev{\mbox{LSTM}}(e_t, \cev{h}_{t+1}) \\
h_t &= [\vec{h}_t, \cev{h}_t ].
\end{split}
\end{equation}
Given $H=\{h_t\}_{t=1}^T$, we apply three trainable vectors $f_m$ where $m\in\{\mbox{subj}, \mbox{loc}, \mbox{rel}\}$, computing the attention on each word~\cite{yang2016hierarchical} for each module:
\begin{equation}
\begin{split}\nonumber
a_{m,t} = \frac{\exp{(f_{m}^T h_t)}}{\sum_{k=1}^T \exp{(f_{m}^T h_k)}}
\end{split}
\end{equation}
The weighted sum of word embeddings is used as the modular phrase embedding: 
\begin{equation}
\begin{split}\nonumber
q^{m}&=\sum_{t=1}^T a_{m,t}e_t\\
\end{split}
\end{equation}

Different from relationship detection~\cite{ronghang16relationship} where phrases are always decomposed as (Subject, Preposition/Verb, Object) triplets, referring expressions have no such well-posed structure.
For example, expressions like ``smiling boy'' only contain language relevant to the subject module, while expressions like ``man on left'' are relevant to the subject and location modules, and ``cat on the chair'' are relevant to the subject and relationship modules.
To handle this variance, we compute 3 module weights for the expression, weighting how much each module contributes to the expression-object score.
We concatenate the first and last hidden vectors from $H$ which memorizes both structure and semantics of the whole expression, then use another fully-connected (FC) layer to transform it into 3 module weights:
\begin{equation}\nonumber
[w_{subj}, w_{loc}, w_{rel}] = \mbox{softmax}(W_{m}^T[h_0, h_T]+b_m)
\end{equation}

\subsection{Visual Modules}
\vspace{-.1cm}
While most previous work~\cite{yu2016refer,yu2016joint,mao2015generation,nagaraja16refexp} evaluates CNN features for each region proposal/candidate object, we use Faster R-CNN~\cite{ren2015faster} as the backbone net for a faster and more principled implementation.
Additionally, we use ResNet~\cite{he2016deep} as our main feature extractor, but also provide comparisons to previous methods using the same VGGNet features~\cite{simonyan2014very} (in Sec.~\ref{sec:results_comprehension}).

Given an image and a set of candidates $o_i$, we run Faster R-CNN to extract their region representations.
Specifically, we forward the whole image into Faster R-CNN and crop the C3 feature (last convolutional output of 3rd-stage) for each $o_i$, following which we further compute the C4 feature (last convolutional output of 4th-stage).
In Faster R-CNN, C4 typically contains higher-level visual cues for category prediction, while C3 contains relatively lower-level cues including colors and shapes for proposal judgment, making both useful for our purposes.
In the end, we compute the matching score for each $o_i$ given each modular phrase embedding, i.e., $S(o_i|q^{subj})$, $S(o_i|q^{loc})$ and $S(o_i|q^{rel})$.

\begin{figure*}[t!]
\centering
\includegraphics[width=0.90\textwidth]{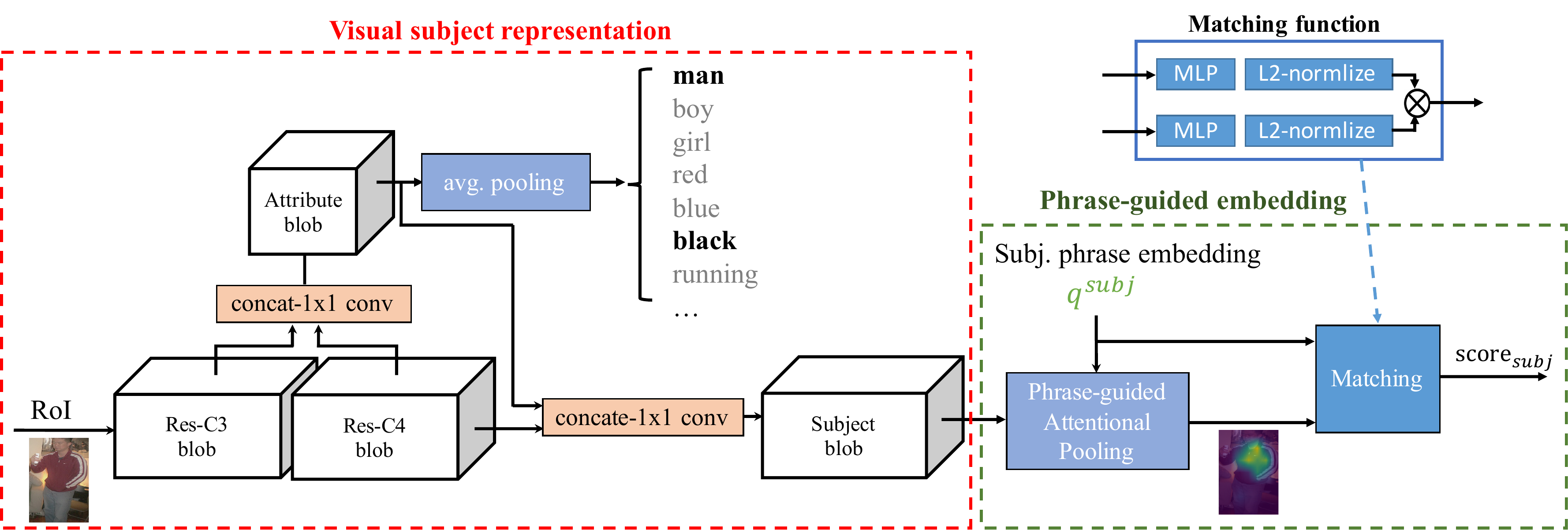}
\caption{The subject module is composed of a visual subject representation and phrase-guided embedding. An attribute prediction branch is added after the ResNet-C4 stage and the 1x1 convolution output of attribute prediction and C4 is used as the subject visual representation. The subject phrase embedding attentively pools over the spatial region and feeds the pooled feature into the matching function.}
\label{fig:subj_module}
\end{figure*}

\subsubsection{Subject Module}
Our subject module is illustrated in Fig.~\ref{fig:subj_module}. Given the C3 and C4 features of a candidate $o_i$, we forward them to two tasks.
The first is attribute prediction, helping produce a representation that can understand appearance characteristics of the candidate.
The second is the phrase-guided attentional pooling to focus on relevant regions within object bounding boxes. 

\textbf{Attribute Prediction}: Attributes are frequently used in referring expressions to differentiate between objects of the same category, e.g. ``woman in red'' or ``the fuzzy cat''.
Inspired by previous work~\cite{yao2016boosting,wu2017image,you2016image,liu2017referring,su2017reasoning}, we add an attribute prediction branch in our subject module.
While preparing the attribute labels in the training set, we first run a template parser~\cite{kazemzadeh2014referitgame} to obtain color and generic attribute words, with low-frequency words removed.
We combine both C3 and C4 for predicting attributes as both low and high-level visual cues are important. The concatenation of C3 and C4 is followed with a $1\times1$ convolution to produce an attribute feature blob.
After average pooling, we get the attribute representation of the candidate region.
A binary cross-entropy loss is used for multi-attribute classification:
\begin{equation}
\nonumber
L_{subj}^{attr} = \lambda_{attr}\sum_i \sum_j w^{attr}_j [\mbox{log}(p_{ij}) + (1-y_{ij})\mbox{log}(1-p_{ij})]
\end{equation}
where $w^{attr}_j=1/\sqrt{\mbox{freq}_{attr}}$ weights the attribute labels, easing unbalanced data issues. During training, only expressions with attribute words go through this branch.

\textbf{Phrase-guided Attentional Pooling}: 
The subject description varies depending on what information is most salient about the object. Take people for example. Sometimes a person is described by their accessories, e.g., ``girl in glasses''; or sometimes particular clothing items may be mentioned, e.g., ``woman in white pants''. Thus, we allow our subject module to localize relevant regions within a bounding box through ``in-box'' attention.
To compute spatial attention, we first concatenate the attribute blob and C4, then use a $1\times1$ convolution to fuse them into a subject blob, which consists of spatial grid of features $V\in R^{d\times G}$, where $G=14\times 14$.
Given the subject phrase embedding $q^{subj}$, we compute its attention on each grid location:
\begin{equation}
\nonumber
\begin{split}
H_a &= \mbox{tanh}(W_v V + W_q q^{subj})\\
a^v &= \mbox{softmax}(w_{h,a}^T H_a) \\
\end{split}
\end{equation}
The weighted sum of $V$ is the final subject visual representation for the candidate region $o_i$:
\begin{equation}\nonumber
\widetilde{v}^{subj}_i = \sum_{i=1}^{G} a^v_i v_i
\end{equation}

\textbf{Matching Function}:
We measure the similarity between the subject representation $\widetilde{v}^{subj}_i$ and phrase embedding $q_{subj}$ using a matching function, i.e, $S(o_i|q^{subj})=F(\widetilde{v}^{subj}_i, q^{subj})$.
As shown in top-right of Fig.~\ref{fig:subj_module}, it consists of two MLPs (multi-layer perceptions) and two L2 normalization layers following each input.
Each MLP is composed of two fully connected layers with ReLU activations, serving to transform the visual and phrase representation into a common embedding space.
The inner-product of the two l2-normalized representations is computed as their similarity score. The same matching function is used to compute the location score $S(o_i|q^{loc})$, and relationship score $S(o_i|q^{rel})$.

\subsubsection{Location Module}\label{sec:loc_module}
\begin{figure}[h]
\centering
\includegraphics[width=0.45\textwidth]{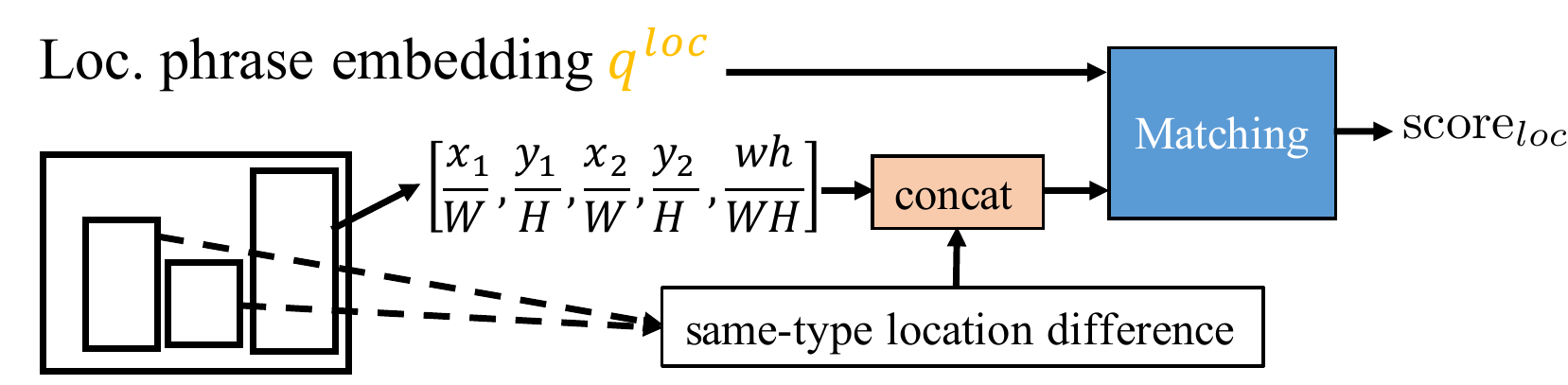}\\
\caption{Location Module}
\label{fig:loc_module}
\end{figure}
Our location module is shown in Fig.~\ref{fig:loc_module}.
Location is frequently used in referring expressions with
about 41\% expressions from RefCOCO and 36\% expressions from RefCOCOg containing absolute location words~\cite{kazemzadeh2014referitgame}, e.g.
``cat on the right'' indicating the object location in the image.
Following previous work~\cite{yu2016refer}\cite{yu2016joint}, we use a 5-d vector $l_i$ to encode the top-left position, bottom-right position and relative area to the image for the candidate object, i.e., $l_i=[\frac{x_{tl}}{W}, \frac{y_{tl}}{H}, \frac{x_{br}}{W}, \frac{y_{br}}{H}, \frac{w\cdot h}{W\cdot H}]$.

Additionally, expressions like ``dog in the middle'' and ``second left person'' imply relative positioning among objects of the same category.
We encode the relative location representation of a candidate object by choosing up to five surrounding objects of the same category and calculating their offsets and area ratio, i.e., $\delta l_{ij}=[\frac{[\bigtriangleup x_{tl}]_{ij}}{w_i}, \frac{[\bigtriangleup y_{tl}]_{ij}}{h_i}, \frac{[\bigtriangleup x_{br}]_{ij}}{w_i}, \frac{[\bigtriangleup y_{br}]_{ij}}{h_i}, \frac{w_j h_j}{w_i h_i}]$. 
The final location representation for the target object is:
\begin{equation}\nonumber
\widetilde{l}^{loc}_i = W_l [l_i; \delta l_i] + b_l
\end{equation}
and the location module matching score between $o_i$ and $q^{loc}$ is $S(o_i|q^{loc})=F(\widetilde{l}^{loc}_i, q^{loc})$.

\subsubsection{Relationship Module}\label{sec:rel_module}
\begin{figure}[h]
\centering
\includegraphics[width=0.45\textwidth]{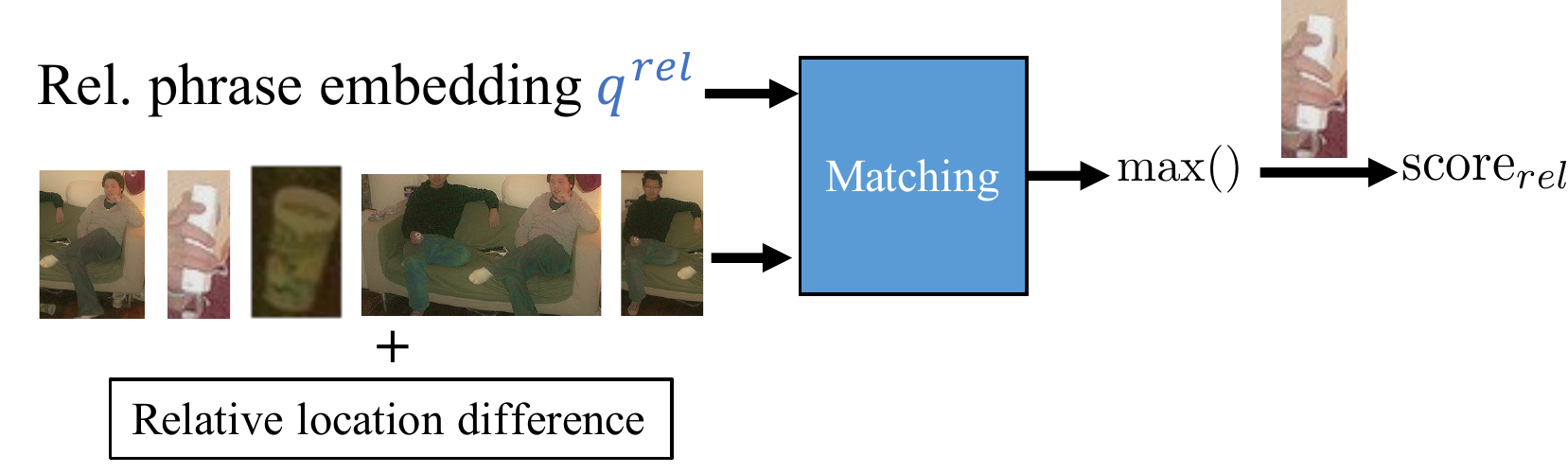}\\
\caption{Relationship Module}
\label{fig:rel_module}
\end{figure}
While the subject module deals with ``in-box'' details about the target object, some other expressions may involve its relationship with other ``out-of-box'' objects, e.g., ``cat on chaise lounge''.
The relationship module is used to address these cases.
As in Fig.~\ref{fig:rel_module}, given a candidate object $o_i$ we first look for its surrounding (up-to-five) objects $o_{ij}$ regardless of their categories.
We use the average-pooled C4 feature as the appearance feature $v_{ij}$ of each supporting object.
Then, we encode their offsets to the candidate object via $\delta m_{ij}=[\frac{[\bigtriangleup x_{tl}]_{ij}}{w_i}, \frac{[\bigtriangleup y_{tl}]_{ij}}{h_i}, \frac{[\bigtriangleup x_{br}]_{ij}}{w_i}, \frac{[\bigtriangleup y_{br}]_{ij}}{h_i}, \frac{w_j h_j}{w_i h_i}]$.
The visual representation for each surrounding object is then:
\begin{equation}\nonumber
\widetilde{v}^{rel}_{ij} = W_r [v_{ij}; \delta m_{ij}] + b_r
\end{equation}
We compute the matching score for each of them with $q^{rel}$ and pick the highest one as the relationship score, i.e.,
\begin{equation}\nonumber
S(o_i|q^{rel})=\mbox{max}_{j\neq i}F(\widetilde{v}^{rel}_{ij}, q^{rel})
\end{equation}
This can be regarded as weakly-supervised Multiple Instance Learning (MIL) which is similar to~\cite{ronghang16relationship}\cite{nagaraja16refexp}. 

\subsection{Loss Function}\label{sec:loss}
\vspace{-.1cm}
The overall weighted matching score for candidate object $o_i$ and expression $r$ is:
\begin{equation}\label{eqn:score}
\resizebox{.43\textwidth}{!}{
$S(o_i|r) = w_{subj}S(o_i|q^{subj}) + w_{loc}S(o_i|q^{loc}) + w_{rel}S(o_i|q^{rel})$
}
\end{equation}

During training, for each given positive pair of $(o_i, r_i)$, we randomly sample two negative pairs $(o_i, r_j)$ and $(o_k, r_i)$, where $r_j$ is the expression describing some other object and $o_k$ is some other object in the same image, to calculate a combined hinge loss,
\begin{equation}\nonumber
\small
\begin{split}
L_{rank}=\sum_i[&\lambda_1 \mbox{max}(0, \Delta+S(o_i|r_j)-S(o_i|r_i)) \\
+&\lambda_2 \mbox{max}(0, \Delta+S(o_k|r_i)-S(o_i|r_i))]
\end{split}
\end{equation}
The overall loss incorporates both attributes cross-entropy loss and ranking loss: $
L=L_{subj}^{attr}+L_{rank}$.


\section{Experiments}

\begin{table*}[t]
\footnotesize
\begin{center}
\resizebox{2.0\columnwidth}{!}{%
\begin{tabular}{| c | l | l | c | c | c || c | c | c || c || c | c |}
\hline
&&& \multicolumn{3}{c}{RefCOCO} & \multicolumn{3}{|c|}{RefCOCO+} & \multicolumn{3}{|c|}{RefCOCOg}\\
\cline{2-12}
&&feature& val & testA & testB & val & testA & testB & val* & val & test\\
\hline\hline
1 & Mao~\cite{mao2015generation}       & vgg16 & - & 63.15 & 64.21 & - & 48.73 & 42.13 & 62.14 & - & - \\
2 & Varun~\cite{nagaraja16refexp}      & vgg16 & 76.90 & 75.60 & 78.00 & - & - & - & - & - & 68.40 \\ 
3 & Luo~\cite{luo2017comprehension}    & vgg16 & - & 74.04 & 73.43 & - & 60.26 & 55.03 & 65.36 & - & - \\
4 & CMN~\cite{ronghang16relationship}  & vgg16-frcn & - & - & - & - & - & - & 69.30 & - & - \\
5 & Speaker/visdif~\cite{yu2016refer}  & vgg16 & 76.18 & 74.39 & 77.30 & 58.94 & 61.29 & 56.24 & 59.40 & - & - \\
6 & Listener~\cite{yu2016joint}        & vgg16 & 77.48 & 76.58 & 78.94 & 60.50 & 61.39 & 58.11 & 71.12 & 69.93 & 69.03 \\
7 & \textbf{Speaker}+Listener+Reinforcer~\cite{yu2016joint}& vgg16 & 79.56 &  78.95 & 80.22 & 62.26 & 64.60 & 59.62 & 72.63 & 71.65 & 71.92 \\
8 & Speaker+\textbf{Listener}+Reinforcer~\cite{yu2016joint}& vgg16 & 78.36 & 77.97 & 79.86 & 61.33 & 63.10 & 58.19 & 72.02 & 71.32 & 71.72\\
\hline
9 & MAttN:subj(+attr)+loc(+dif)+rel & vgg16 & 80.94 & 79.99 & 82.30 & 63.07 & 65.04 & 61.77 & 73.08 & 73.04 & 72.79 \\
10 & MAttN:subj(+attr)+loc(+dif)+rel & res101-frcn & 83.54 & 82.66 & 84.17 & 68.34 & 69.93 & 65.90 & - & 76.63 & 75.92 \\ 
11& MAttN:subj(+attr+attn)+loc(+dif)+rel & res101-frcn & \textbf{85.65} & \textbf{85.26} & \textbf{84.57} & \textbf{71.01} & \textbf{75.13} & \textbf{66.17} & - & \textbf{78.10} & \textbf{78.12} \\ 
\hline
\end{tabular}
}
\end{center}
\vspace{-10pt}
\caption{Comparison with state-of-the-art approaches on ground-truth MS COCO regions.}
\label{table:comprehension_comparison}
\end{table*}

\begin{table*}[t]
\footnotesize
\begin{center}
\resizebox{2.0\columnwidth}{!}{%
\begin{tabular}{| c | l | c | c | c || c | c | c || c | c |}
\hline
&& \multicolumn{3}{c}{RefCOCO} & \multicolumn{3}{|c|}{RefCOCO+} & \multicolumn{2}{|c|}{RefCOCOg}\\
\cline{2-10}
&& val & testA & testB & val & testA & testB & val & test\\
\hline\hline
1 & Matching:subj+loc                 & 79.14 & 79.42 & 80.42 & 62.17 & 63.53 & 59.87 & 70.45 & 70.92 \\
2 & MAttN:subj+loc                      & 79.68 & 80.20 & 81.49 & 62.71 & 64.20 & 60.65 & 72.12 & 72.62 \\
3 & MAttN:subj+loc(+dif)                & 82.06 & 81.28 & 83.20 & 64.84 & 65.77 & 64.55 & 75.33 & 74.46 \\ 
4 & MAttN:subj+loc(+dif)+rel            & 82.54 & 81.58 & 83.34 & 65.84 & 66.59 & 65.08 & 75.96 & 74.56 \\
5 & MAttN:subj(+attr)+loc(+dif)+rel      & 83.54 & 82.66 & 84.17 & 68.34 & 69.93 & 65.90 & 76.63 & 75.92 \\ 
6 & MAttN:subj(+attr+attn)+loc(+dif)+rel & \textbf{85.65} & \textbf{85.26} & \textbf{84.57} & \textbf{71.01} & \textbf{75.13} & \textbf{66.17} & \textbf{78.10} & \textbf{78.12} \\ 
\hline
7 & parser+MAttN:subj(+attr+attn)+loc(+dif)+rel& 80.20 & 79.10 & 81.22 & 66.08 & 68.30 & 62.94 & 73.82 & 73.72 \\
\hline
\end{tabular}
}
\end{center}
\vspace{-10pt}
\caption{Ablation study of MAttNet using different combination of modules. The feature used here is res101-frcn.}
\label{table:comprehension_ablation}
\end{table*}

\begin{table*}[t]
\footnotesize
\begin{center}
\resizebox{2.0\columnwidth}{!}{%
\begin{tabular}{| c | l | c | c | c | c || c | c | c || c | c |}
\hline
&&& \multicolumn{3}{c}{RefCOCO} & \multicolumn{3}{|c|}{RefCOCO+} & \multicolumn{2}{|c|}{RefCOCOg}\\
\cline{2-11}
&&detector& val & testA & testB & val & testA & testB & val & test\\
\hline\hline
1 & \textbf{Speaker}+Listener+Reinforcer~\cite{yu2016joint}  & res101-frcn & 69.48 & 73.71 & 64.96 & 55.71 & 60.74 & 48.80 & 60.21 & 59.63\\
2 & Speaker+\textbf{Listener}+Reinforcer~\cite{yu2016joint}  & res101-frcn & 68.95 & 73.10 & 64.85 & 54.89 & 60.04 & 49.56 & 59.33 & 59.21\\
\hline
3 & Matching:subj+loc                 & res101-frcn & 72.28 & 75.43 & 67.87 & 58.42 & 61.46 & 52.73 & 64.15 & 63.25 \\
4 & MAttN:subj+loc                     & res101-frcn & 72.72 & 76.17 & 68.18 & 58.70 & 61.65 & 53.41 & 64.40 & 63.74 \\
5 & MAttN:subj+loc(+dif)               & res101-frcn & 72.96 & 76.61 & 68.20 & 58.91 & 63.06 & 55.19 & 64.66 & 63.88 \\ 
6 & MAttN:subj+loc(+dif)+rel           & res101-frcn & 73.25 & 76.77 & 68.44 & 59.45 & 63.31 & 55.68 & 64.87 & 64.01 \\
7 & MAttN:subj(+attr)+loc(+dif)+rel    & res101-frcn  & 74.51 & 77.81 & 68.39 & 62.13 & 66.33 & 55.75 & 65.33 & 65.19 \\ 
8 & MAttN:subj(+attr+attn)+loc(+dif)+rel & res101-frcn & 76.40 & 80.43 & 69.28 & 64.93 & 70.26 & 56.00 & \textbf{66.67} & 67.01 \\ \hline
9 & MAttN:subj(+attr+attn)+loc(+dif)+rel & res101-mrcn & \textbf{76.65} & \textbf{81.14} & \textbf{69.99} & \textbf{65.33} & \textbf{71.62} & \textbf{56.02} & 66.58 & \textbf{67.27} \\
\hline
\end{tabular}
}
\end{center}
\vspace{-10pt}
\caption{Ablation study of MAttNet on fully-automatic comprehension task using different combination of modules. The features used here are res101-frcn, except the last row using res101-mrcn.}
\label{table:comprehension_automatic}
\end{table*}

\begin{table*}[t]
\footnotesize
\begin{center}
\resizebox{1.8\columnwidth}{!}{%
\begin{tabular}{ c | c | c | c  c  c  c  c  c }
\multicolumn{9}{c}{RefCOCO}\\
\hline
Model & Backbone Net & Split & Pr@0.5 & Pr@0.6 & Pr@0.7 & Pr@0.8 & Pr@0.9 & IoU \\
\hline
D+RMI+DCRF~\cite{liu2017recurrent}    & res101-DeepLab & val & 42.99 & 33.24 & 22.75 & 12.11 & 2.23 & 45.18 \\
MAttNet                                & res101-mrcn & val & \textbf{75.16} & \textbf{72.55} & \textbf{67.83} &
\textbf{54.79} & \textbf{16.81} & \textbf{56.51} \\
\hline
D+RMI+DCRF~\cite{liu2017recurrent}    & res101-DeepLab & testA & 42.99 & 33.59 & 23.69 & 12.94 & 2.44 & 45.69 \\
MAttNet                                & res101-mrcn & testA & \textbf{79.55} & \textbf{77.60} & \textbf{72.53} & \textbf{59.01} & \textbf{13.79} & \textbf{62.37} \\
\hline
D+RMI+DCRF~\cite{liu2017recurrent}    & res101-DeepLab & testB & 44.99 & 32.21 & 22.69 & 11.84 & 2.65 & 45.57\\
MAttNet                                & res101-mrcn & testB & \textbf{68.87} & \textbf{65.06} & \textbf{60.02} & \textbf{48.91} & \textbf{21.37} & \textbf{51.70} \\
\hline
\end{tabular}
}
\end{center}

\begin{center}
\resizebox{1.8\columnwidth}{!}{%
\begin{tabular}{ c | c | c | c  c  c  c  c  c }
\multicolumn{9}{c}{RefCOCO+}\\
\hline
Model & Backbone Net & Split & Pr@0.5 & Pr@0.6 & Pr@0.7 & Pr@0.8 & Pr@0.9 & IoU \\
\hline
D+RMI+DCRF~\cite{liu2017recurrent}   & res101-DeepLab & val & 20.52 & 14.02 & 8.46 & 3.77 & 0.62 & 29.86 \\
MAttNet                               & res101-mrcn & val & \textbf{64.11} & \textbf{61.87} & \textbf{58.06} & \textbf{47.42} & \textbf{14.16} & \textbf{46.67} \\
\hline
D+RMI+DCRF~\cite{liu2017recurrent}   & res101-DeepLab & testA & 21.22 & 14.43 & 8.99 & 3.91 & 0.49 & 30.48 \\
MAttNet                               & res101-mrcn & testA & \textbf{70.12} & \textbf{68.48} & \textbf{63.97} & \textbf{52.13} & \textbf{12.28} & \textbf{52.39}\\
\hline
D+RMI+DCRF~\cite{liu2017recurrent}   & res101-DeepLab & testB & 20.78 & 14.56 & 8.80 & 4.58 & 0.80 & 29.50\\
MAttNet                              & res101-mrcn & testB & \textbf{54.82} & \textbf{51.73} & \textbf{47.27} & \textbf{38.58} & \textbf{17.00} & \textbf{40.08} \\
\hline
\end{tabular}
}
\end{center}

\begin{center}
\resizebox{1.8\columnwidth}{!}{%
\begin{tabular}{ c | c | c | c  c  c  c  c  c }
\multicolumn{9}{c}{RefCOCOg}\\
\hline
Model & Backbone Net & Split & Pr@0.5 & Pr@0.6 & Pr@0.7 & Pr@0.8 & Pr@0.9 & IoU \\
\hline
MAttNet & res101-mrcn & val  & 64.48 & 61.52 & 56.50 & 43.97 & 14.67 & 47.64 \\
\hline
MAttNet & res101-mrcn & test & 65.60 & 62.92 & 57.31 & 44.44 & 12.55 & 48.61 \\
\hline
\end{tabular}
}
\end{center}
\vspace{-10pt}
\caption{Comparison of segmentation performance on RefCOCO, RefCOCO+, and our results on RefCOCOg.}
\label{table:comprehension_segmentation}
\end{table*}

\begin{figure*}[t]
\includegraphics[width=0.92\textwidth]{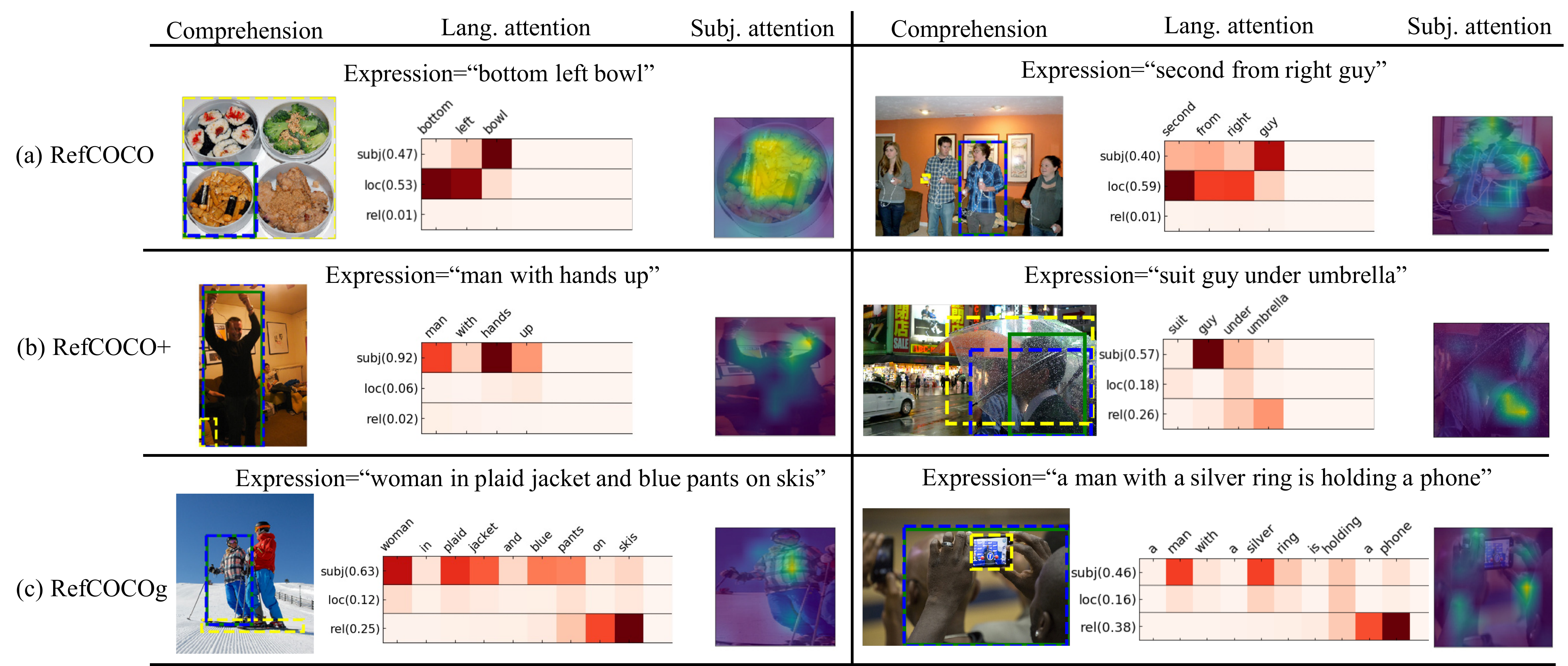}
\caption{Examples of fully automatic comprehension. The blue dotted boxes show our prediction with the relative regions in yellow dotted boxes, and the green boxes are the ground-truth. The word attention is multiplied by module weight.}
\label{fig:examples_correct}
\end{figure*}

\begin{figure}[t]
\includegraphics[width=0.45\textwidth]{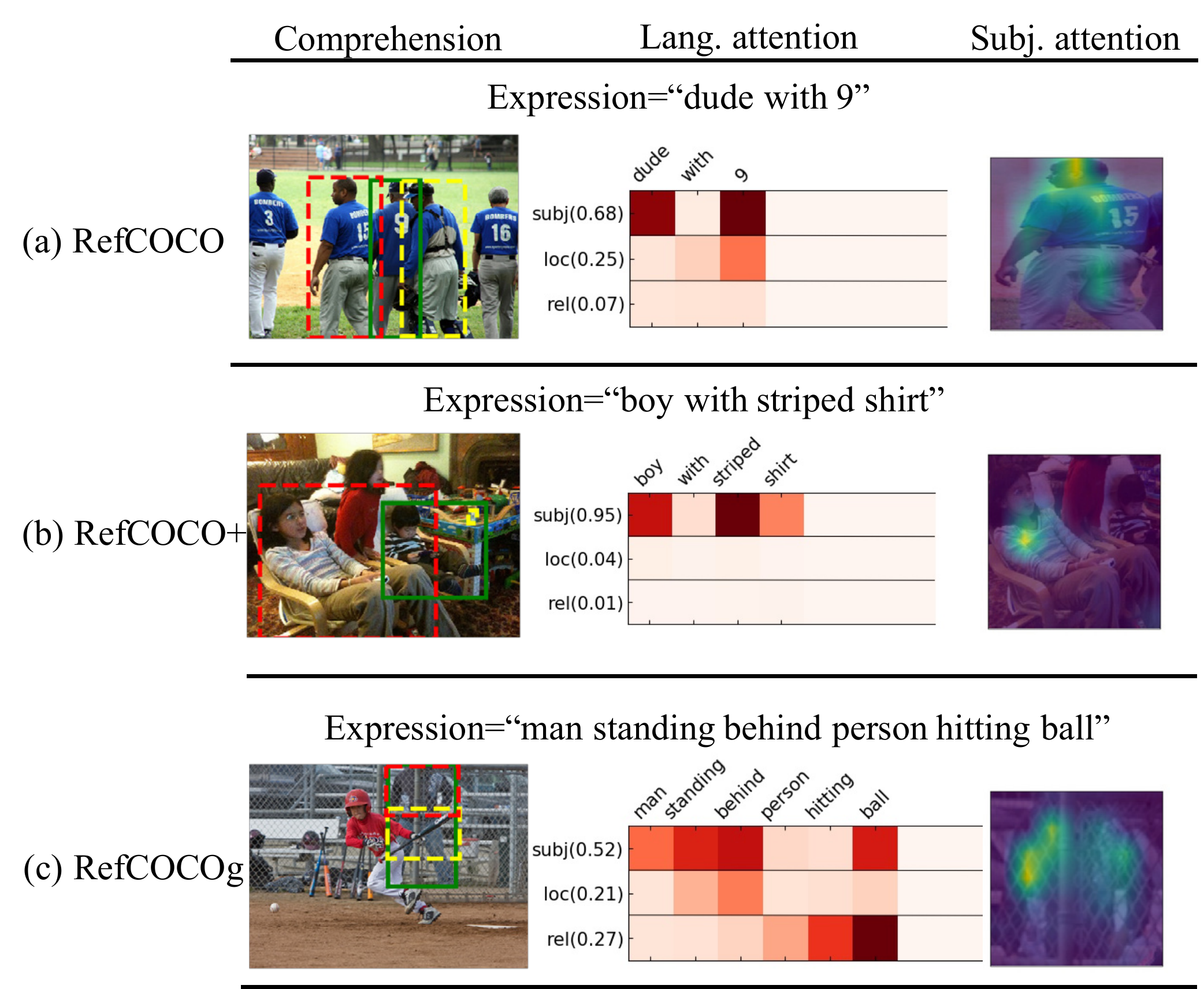}
\caption{Examples of incorrect comprehensions. 
Red dotted boxes show our wrong prediction.
}
\label{fig:examples_wrong}
\end{figure}

\begin{figure}[t]
\includegraphics[width=0.45\textwidth]{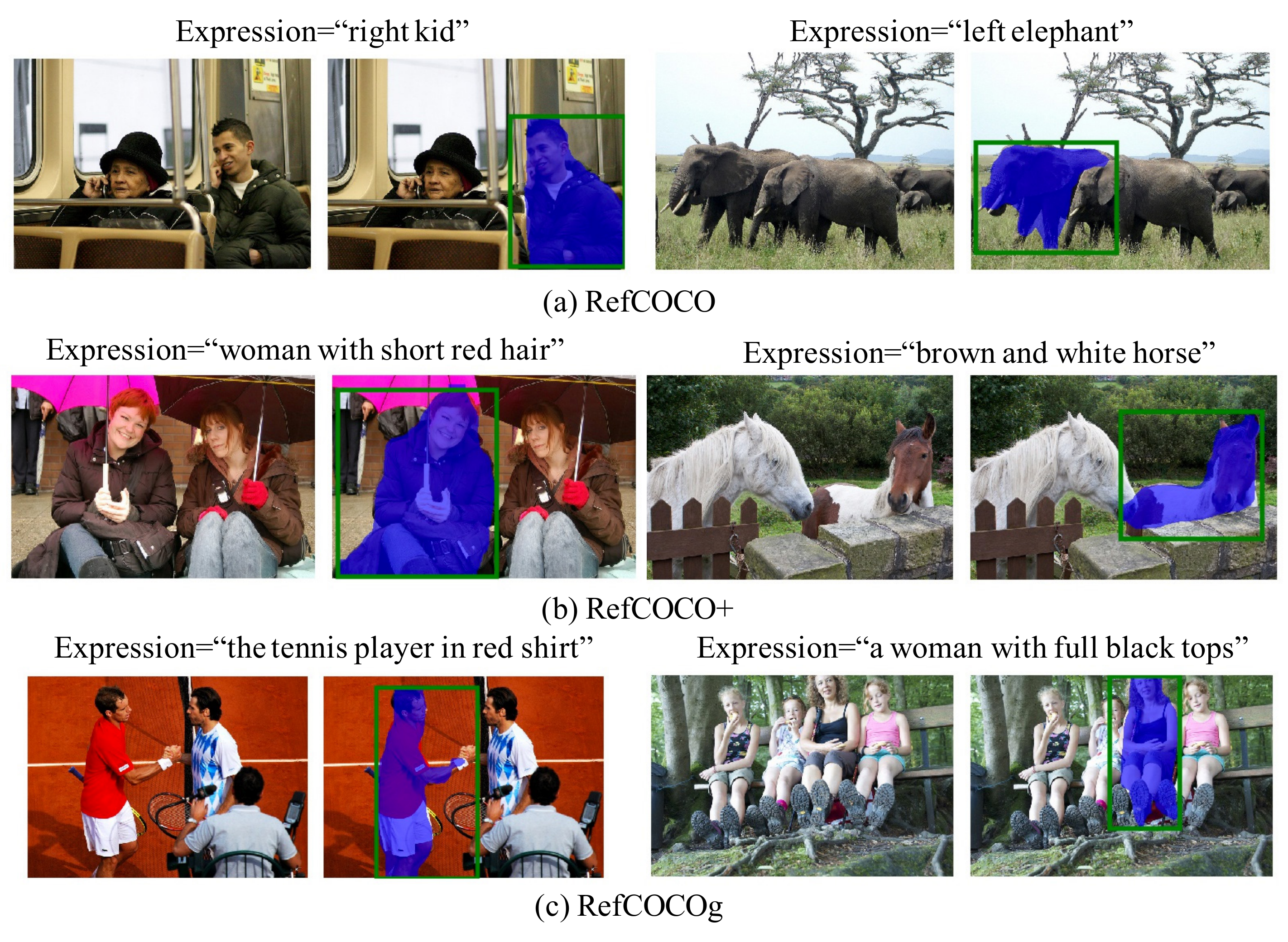}
\caption{Examples of fully-automatic MAttNet referential segmentation. 
}
\label{fig:examples_segmentation}
\end{figure}

\subsection{Datasets}
\vspace{-.1cm}
We use 3 referring expression datasets: RefCOCO, RefCOCO+~\cite{kazemzadeh2014referitgame}, and RefCOCOg~\cite{mao2015generation} for evaluation, all collected on MS COCO images~\cite{lin2014microsoft}, but with several differences.
1) RefCOCO and RefCOCO+ were collected in an interactive game interface, while RefCOCOg was collected in a non-interactive setting thereby producing longer expressions, 3.5 and 8.4 words on average respectively.
2) RefCOCO and RefCOCO+ contain more same-type objects, 3.9 vs 1.63 respectively.
3) RefCOCO+ forbids using absolute location words, making the data more focused on appearance differentiators.

During testing, RefCOCO and RefCOCO+ provide person vs. object splits for evaluation, where images containing multiple people are in ``testA'' and those containing multiple objects of other categories are in ``testB''.
There is no overlap between training, validation and testing images.
RefCOCOg has two types of data partitions.
The first~\cite{mao2015generation} divides the dataset by randomly partitioning objects into training and validation splits.
As the testing split has not been released, most recent work evaluates performance on the validation set.
We denote this validation split as RefCOCOg's ``val*''.
Note, since this data is split by objects the same image could appear in both training and validation.
The second partition~\cite{nagaraja16refexp} is composed by randomly partitioning images into training, validation and testing splits.
We denote its validation and testing splits as RefCOCOg's ``val'' and ``test'', and run most experiments on this split.

\subsection{Results: Referring Expression Comprehension}
\label{sec:results_comprehension}
\vspace{-.1cm}
Given a test image, $I$, with a set of proposals/objects, $O=\{o_i\}_{i=1}^N$, we use Eqn.~\ref{eqn:score} to compute the matching score $S(o_i|r)$ for each proposal/object given the input expression $r$, and pick the one with the highest score.
For evaluation, we compute the intersection-over-union (IoU) of the selected region with the ground-truth bounding box, considering IoU $> 0.5$ a correct comprehension.

First, we compare our model with previous methods using COCO's ground-truth object bounding boxes as proposals. Results are shown in Table.~\ref{table:comprehension_comparison}.
As all of the previous methods (Line 1-8) used a 16-layer VGGNet (vgg16) as the feature extractor, we run our experiments using the same feature for fair comparison. Note the flat fc7 is a single 4096-dimensional feature which prevents us from using the phrase-guided attentional pooling in Fig.~\ref{fig:subj_module}, so we use average pooling for subject matching. Despite this, our results (Line 9) still outperform all previous state-of-the-art methods.
After switching to the res101-based Faster R-CNN (res101-frcn) representation, the comprehension accuracy further improves another ${\sim}3\%$ (Line 10).
Note our Faster R-CNN is pre-trained on COCO's training images, excluding those in RefCOCO, RefCOCO+, and RefCOCOg's validation+testing.
Thus no training images are seen during our evaluation\footnote{Such constraint forbids us to evaluate on RefCOCOg's val* using the res101-frcn feature in Table~\ref{table:comprehension_comparison}.}.
Our full model (Line 11) with phrase-guided attentional pooling achieves the highest accuracy over all others by a large margin.

Second, we study the benefits of each module of MAttNet by running ablation experiments (Table.~\ref{table:comprehension_ablation}) with the same res101-frcn features.
As a baseline, we use the concatenation of the regional visual feature and the location feature as the visual representation and the last hidden output of LSTM-encoded expression as the language representation, then feed them into the matching function to obtain the similarity score (Line 1).
Compared with this, a simple two-module MAttNet using the same features (Line 2) already outperforms the baseline, showing the advantage of modular learning.
Line 3 shows the benefit of encoding location (Sec.~\ref{sec:loc_module}).
After adding the relationship module, the performance further improves (Line 4).
Lines 5 and Line 6 show the benefits brought by the attribute sub-branch and the phrase-guided attentional pooling in our subject module.
We find the attentional pooling (Line 6) greatly improves on the person category (testA of RefCOCO and RefCOCO+), demonstrating the advantage of modular attention on understanding localized details like ``girl with red hat''.

Third, we tried training our model using 3 hard-coded phrases from a template language parser~\cite{kazemzadeh2014referitgame}, shown in Line 7 of Table.~\ref{table:comprehension_ablation}, which is ${\sim}5\%$ lower than our end-to-end model (Line 6).
The main reason for this drop is errors made by the external parser which is not tuned for referring expressions.

Fourth, we show results using automatically detected objects from Faster R-CNN, providing an analysis of fully automatic comprehension performance.
Table.~\ref{table:comprehension_automatic} shows the ablation study of fully-automatic MAttNet.
While performance drops due to detection errors, the overall improvements brought by each module are consistent with Table.~\ref{table:comprehension_ablation}, showing the robustness of MAttNet.
Our results also outperform the state-of-the-art~\cite{yu2016joint} (Line 1,2) with a big margin.
Besides, we show the performance when using the detector branch of Mask R-CNN~\cite{he2017mask} (res101-mrcn) in Line 9, whose results are even better than using Faster R-CNN. 

Finally, we show some example visualizations of comprehension using our full model in Fig.~\ref{fig:examples_correct} as well as visualizations of the attention predictions. We observe that our language model is able to attend to the right words for each module even though it is learned in a weakly-supervised manner.
We also observe the expressions in RefCOCO and RefCOCO+ describe the location or details of the target object more frequently while RefCOCOg mentions the relationship between target object and its surrounding object more frequently, which accords with the dataset property.
Note that for some complex expressions like ``woman in plaid jacket and blue pants on skis'' which contains several relationships (last row in Fig.~\ref{fig:examples_correct}), our language model is able to attend to the portion that should be used by the ``in-box'' subject module and the portion that should be used by the ``out-of-box'' relationship module.
Additionally our subject module also displays reasonable spatial ``in-box'' attention, which qualitatively explains why attentional pooling (Table.~\ref{table:comprehension_ablation} Line 6) outperforms average pooling (Table.~\ref{table:comprehension_ablation} Line 5).
For comparison, some incorrect comprehension are shown in Fig.~\ref{fig:examples_wrong}.
Most errors are due to sparsity in the training data, ambiguous expressions, or detection error.

\vspace{-.2cm}
\subsection{Segmentation from Referring Expression}\label{sec:segmentation}
\vspace{-.2cm}
Our model can also be used to address referential object segmentation~\cite{hu2016segmentation,liu2017recurrent}.
Instead of using Faster R-CNN as the backbone net, we now turn to res101-based Mask R-CNN~\cite{he2017mask} (res101-mrcn).
We apply the same procedure described in Sec.~\ref{sec:model} on the detected objects, and use the one with highest matching score as our prediction.
Then we feed the predicted bounding box to the mask branch to obtain a pixel-wise segmentation.
We evaluate the full model of MAttNet and compare with the best results reported in~\cite{liu2017recurrent}.
We use Precision@X ($\mbox{X}\in\{0.5, 0.6, 0.7, 0.8, 0.9\}$)\footnote{Precision@0.5 is the percentage of expressions where the IoU of the predicted segmentation and ground-truth is at least 0.5.} and overall Intersection-over-Union (IoU) as metrics.
Results are shown in Table.~\ref{table:comprehension_segmentation} with our model outperforming state-of-the-art results by a large margin under all metrics\footnote{There is no experiments on RefCOCOg's val/test splits in~\cite{liu2017recurrent}, so we show our performance only for reference in Table~\ref{table:comprehension_segmentation}.}.
As both~\cite{liu2017recurrent} and MAttNet use res101 features, such big gains may be due to our proposed model.
We believe decoupling box localization (comprehension) and segmentation brings a large gain over FCN-style~\cite{long2015fully} foreground/background mask classification~\cite{hu2016segmentation,liu2017recurrent} for this instance-level segmentation problem, but a more end-to-end segmentation system may be studied in future work.
Some referential segmentation examples are shown in Fig.~\ref{fig:examples_segmentation}.

\vspace{-.2cm}
\section{Conclusion}
\vspace{-.25cm}
Our modular attention network addresses variance in referring expressions by attending to both relevant words and visual regions in a modular framework, and dynamically computing an overall matching score. We demonstrate our model's effectiveness on bounding-box-level and pixel-level comprehension, significantly outperforming state-of-the-art.

\smallskip
\noindent
{\bf Acknowledgements:} This research is 
supported by NSF Awards \#1405822, 1562098, 1633295, NVidia, Google Research, Microsoft Research and Adobe Research.

\appendix
\section{Appendix}

\subsection{Training Details}
We optimize our model using Adam with an initial learning rate of 0.0004 and with a batch size of 15 images (and all their expressions).
The learning rate is halved every 8,000 iterations after the first 8,000-iteration warm-up.
The word embedding size and hidden state size of the LSTM are set to 512.
We also set the output of all MLPs and FCs within our model to be 512-dimensional.
To avoid overfitting, we regularize the word-embedding and output layers of the LSTM in the language attention network using dropout with ratio of 0.5.
We also regularize the two inputs (visual and language) of matching function using a dropout with a ratio of 0.2.
For the constrastive pairs, we set $\lambda_1=1.0$ and $\lambda_2=1.0$ in the ranking loss $L_{rank}$. 
Besides, we set $\lambda_{attr}=1.0$ for multi-label attribute cross-entropy loss $L_{subj}^{attr}$.

\subsection{Computational Efficiency}
During training, the full model of MAttNet converges at around 30,000 iterations, which takes around half day using single Titan-X(Pascal).
At inference time, our fully automatic system goes through both Mask R-CNN and MAttNet, which takes on average 0.33 seconds for a forward, where 0.31 seconds are spent on Mask R-CNN and 0.02 seconds on MAttNet.

\begin{figure}[t!]
    \centering
    \begin{subfigure}[t]{0.42\textwidth}
        \raisebox{-\height}{\includegraphics[width=\textwidth]{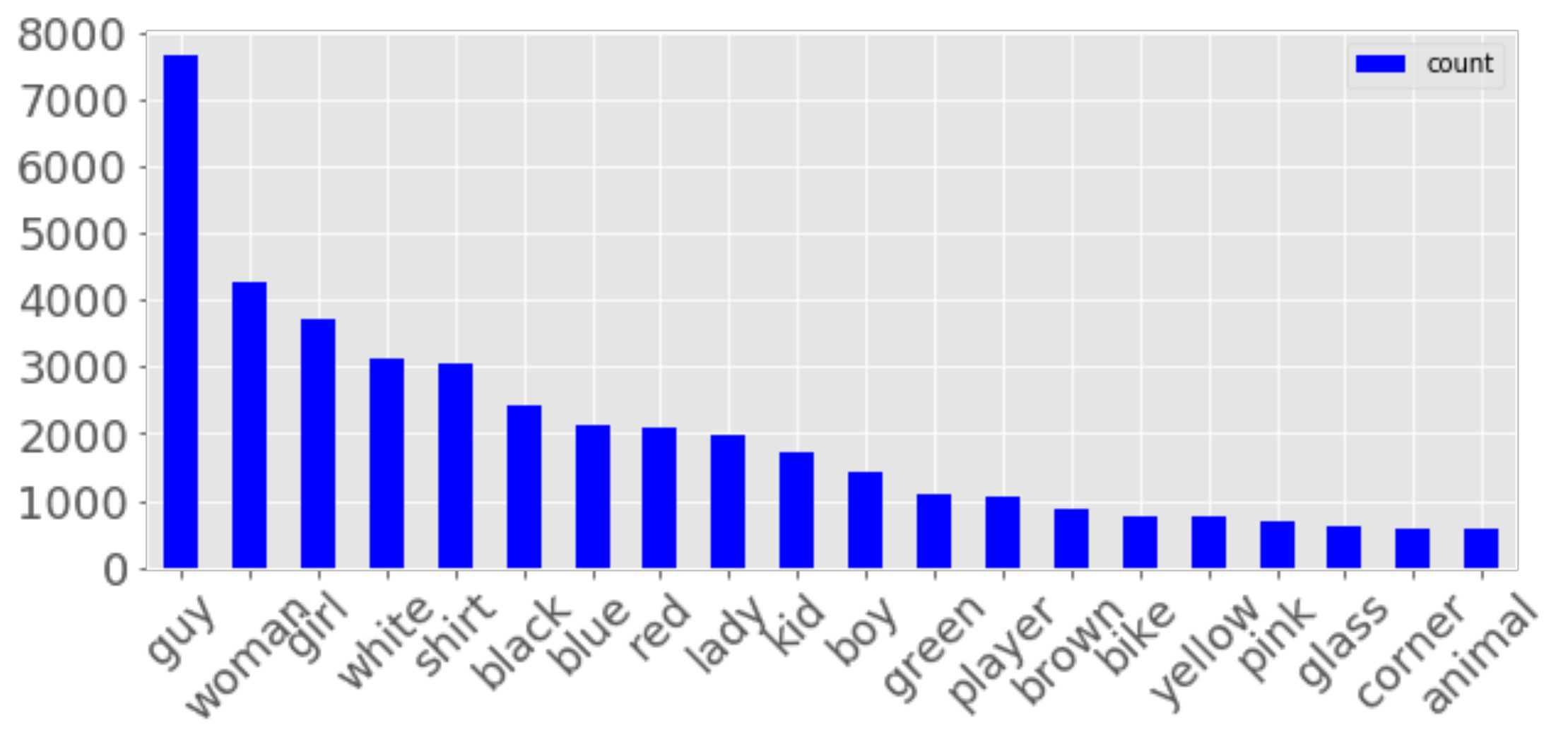}}
        \vspace{-5pt}
        \caption{RefCOCO}
    \end{subfigure}
    \hfill
    \begin{subfigure}[t]{0.42\textwidth}
        \raisebox{-\height}{\includegraphics[width=\textwidth]{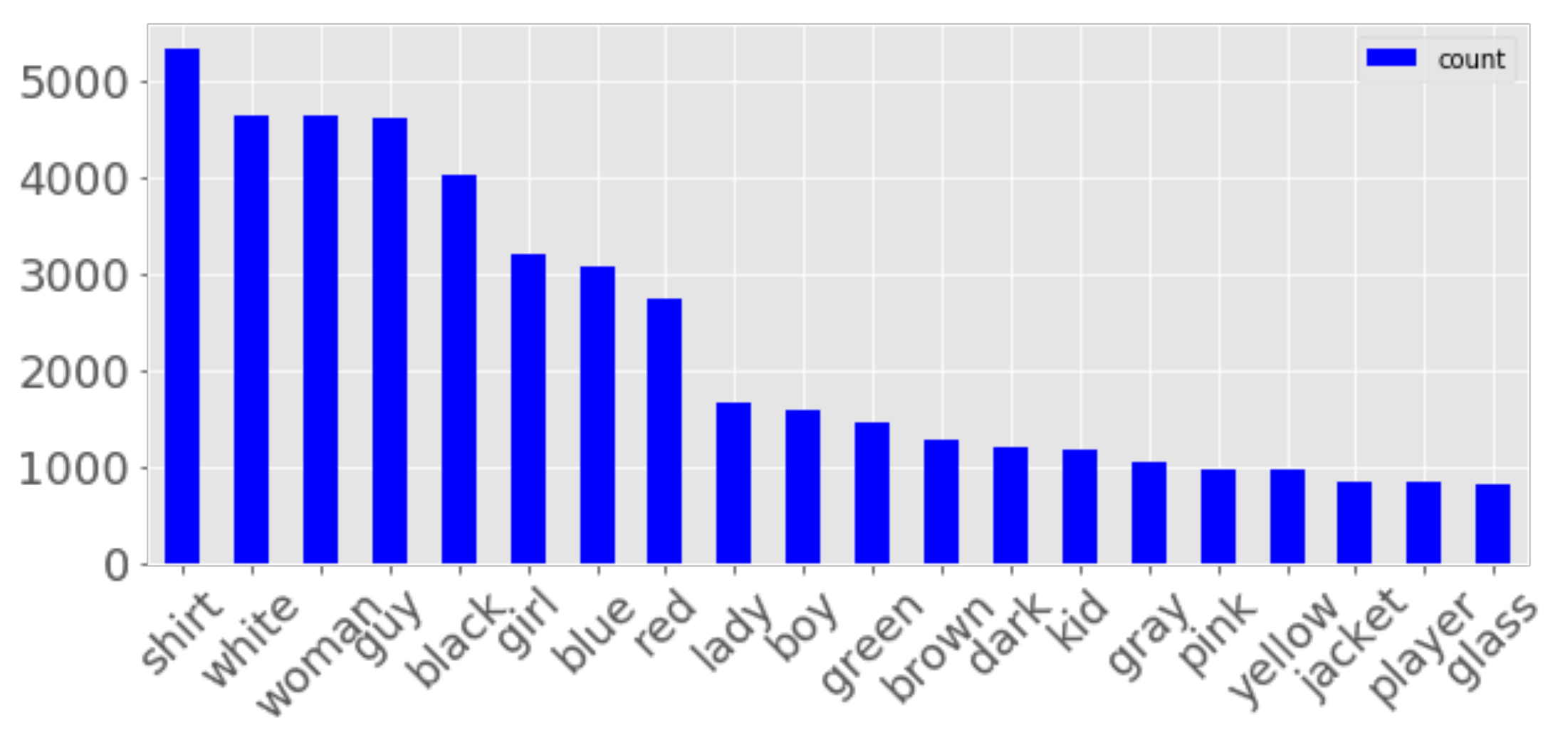}}
        \vspace{-5pt}
        \caption{RefCOCO+}
    \end{subfigure}
    \hfill
    \begin{subfigure}[t]{0.42\textwidth}
        \raisebox{-\height}{\includegraphics[width=\textwidth]{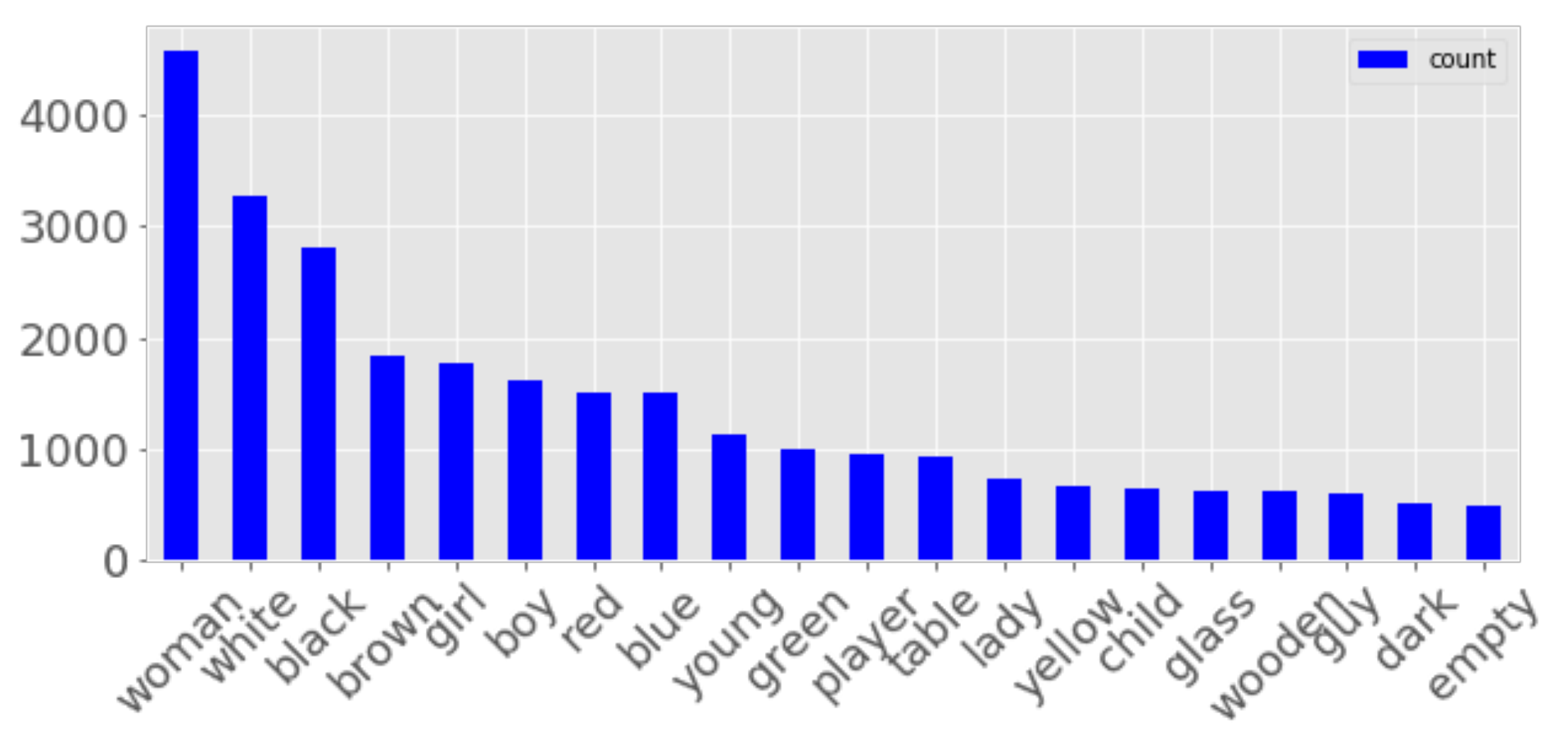}}
        \vspace{-5pt}
        \caption{RefCOCOg}
    \end{subfigure}
    \vspace{-10pt}
    \caption{Attribute histogram for three datasets.}
    \label{fig:word_cloud}
\end{figure}

\subsection{Attribute Prediction}

Our full model is also able to predict attributes during testing.
Our attribute labels are extracted using the template parser~\cite{kazemzadeh2014referitgame}.
We fetch the object name, color and generic attribute words from each expression, with low-frequency words removed.
We use 50 most frequently used attribute words for training.
The histograms for top-20 attribute words are shown in Fig.~\ref{fig:word_cloud}, and the quantitative analysis of our multi-attribute prediction results is shown in Table.~\ref{table:attribute}.

\begin{table}[h]
\footnotesize
\begin{center}
\resizebox{0.8\columnwidth}{!}{%
\begin{tabular}{ l | c |  c  c  c }
 & Split & Precision & Recall & F1 \\ 
\hline
RefCOCO  & val & 63.48 & 29.91 & 40.66 \\
RefCOCO+ & val & 61.78 & 20.11 & 30.14 \\
RefCOCOg & val & 68.18 & 34.79 & 46.07 \\
\hline
\end{tabular}
}
\end{center}
\vspace{-15pt}
\caption{Multi-attribute prediction on the validation split of each dataset.}\label{table:attribute}
\end{table}

\subsection{MAttNet + Grabcut}

\begin{table*}[h!]
\begin{center}
\footnotesize
\resizebox{1.5\columnwidth}{!}{%
\begin{tabular}{ c | c | c  c  c  c  c  c }
\multicolumn{8}{c}{RefCOCO}\\
\hline
Model & Split & Pr@0.5 & Pr@0.6 & Pr@0.7 & Pr@0.8 & Pr@0.9 & IoU \\
\hline
D+RMI+DCRF~\cite{liu2017recurrent}    & val & 42.99 & 33.24 & 22.75 & 12.11 & 2.23 & \textbf{45.18} \\
MAttNet+\textbf{GrabCut}   & val & \textbf{51.25} & \textbf{41.89} & \textbf{29.77} & \textbf{17.13} & \textbf{5.38} & 42.86 \\
\hline
D+RMI+DCRF~\cite{liu2017recurrent}    & testA & 42.99 & 33.59 & 23.69 & 12.94 & 2.44 & \textbf{45.69} \\
MAttNet+\textbf{GrabCut}              & testA & \textbf{52.94} & \textbf{42.60} & \textbf{27.68} & \textbf{13.29} & \textbf{2.92} & 44.37 \\
\hline
D+RMI+DCRF~\cite{liu2017recurrent}    & testB & 44.99 & 32.21 & 22.69 & 11.84 & 2.65 & \textbf{45.57} \\
MAttNet+\textbf{GrabCut}              & testB & \textbf{47.18} & \textbf{38.27} & \textbf{29.97} & \textbf{20.35} & \textbf{7.85} & 40.71 \\
\hline
\end{tabular}
}
\end{center}

\begin{center}
\resizebox{1.5\columnwidth}{!}{%
\begin{tabular}{ c | c | c  c  c  c  c  c }
\multicolumn{8}{c}{RefCOCO+}\\
\hline
Model & Split & Pr@0.5 & Pr@0.6 & Pr@0.7 & Pr@0.8 & Pr@0.9 & IoU \\
\hline
D+RMI+DCRF~\cite{liu2017recurrent}    & val & 20.52 & 14.02 & 8.46 & 3.77 & 0.62 & 29.86 \\
MAttNet+\textbf{GrabCut}   & val & \textbf{45.24} & \textbf{37.09} & \textbf{26.51} & \textbf{14.95} & \textbf{4.34} & \textbf{37.18} \\
\hline
D+RMI+DCRF~\cite{liu2017recurrent}   & testA & 21.22 & 14.43 & 8.99 & 3.91 & 0.49 & 30.48 \\
MAttNet+\textbf{GrabCut}             & testA & \textbf{47.10} & \textbf{37.86} & \textbf{24.66} & \textbf{11.67} & \textbf{2.27} & \textbf{38.32} \\
\hline
D+RMI+DCRF~\cite{liu2017recurrent}   & testB & 20.78 & 14.56 & 8.80 & 4.58 & 0.80 & 29.50\\
MAttNet+\textbf{GrabCut}             & testB & \textbf{38.52} & \textbf{31.13} & \textbf{24.44} & \textbf{16.71} & \textbf{6.20} & \textbf{33.30} \\
\hline
\end{tabular}
}
\end{center}
\vspace{-0.3cm}
    \caption{Comparison of referential segmentation performance between D+RMI+DCRF~\cite{liu2017recurrent} and MatNet+GrabCut.}
    \label{tab:grabcut}
\end{table*}

In Section~\ref{sec:segmentation}, we show MAttNet could be extended to referential segmentation by using Mask R-CNN as the backbone net.
Actually, the mask branch of MAttNet could be any foreground-background decomposition method.
The simplest replacement might be GrabCut.
We show the results of MatNet+GrabCut in Table~\ref{tab:grabcut}.
Note even though GrabCut is an inferior segmentation method, it still far outperforms previous state-of-the-art results~\cite{liu2017recurrent}.
Thus, we believe the way of decoupling box localization (comprehension) and segmentation is more suitable for instance-level referential segmentation task.

\subsection{Mask R-CNN Implementation}
Our implementation of Mask R-CNN\footnote{Our implementation: \href{https://github.com/lichengunc/mask-faster-rcnn}{https://github.com/lichengunc/mask-faster-rcnn}.} is based on the single-GPU Faster R-CNN implementation~\cite{chen2017implementation}.
For the mask branch, we follow the structure in the original paper~\cite{he2017mask}, with several differences:
1) We sample $R=256$ regions from $N=1$ image during each forward-backward propagation due to the constraint of single GPU, while~\cite{he2017mask} samples $R=128$ regions from $N=16$ images using 8 GPUs.
2) During training, the shorter edge of our resized image is 600 pixels instead of 800 pixels, for saving memory.
3) Our model is trained on a union of COCO's 80k train and 35k subset of val (trainval35k) images excluding the val/test (valtest4k) images in RefCOCO, RefCOCO+ and RefCOCOg.

We firstly show the comparison between Faster R-CNN and Mask R-CNN on object detection in Table.~\ref{table:detection}.
Both models are based on ResNet101 and were trained using same setting.
In the main paper, we denote them as res101-frcn and res101-mrcn respectively.
It shows that Mask R-CNN has higher AP than Faster R-CNN due to the multi-task training (with additional mask supervision).

\begin{table}[h]
\begin{center}
\resizebox{0.7\columnwidth}{!}{%
\begin{tabular}{ c | c  c  c}
net & $AP^{bb}$ & $AP_{50}^{bb}$ & $AP_{75}^{bb}$ \\
\hline
res101-frcn  & 34.1 & 53.7 & 36.8  \\
res101-mrcn  & 35.8 & 55.3 & 38.6  \\
\hline
\end{tabular}
}
\end{center}
\vspace{-15pt}
\caption{Object detection results.}\label{table:detection}
\end{table}

\begin{table}[h]
\begin{center}
\resizebox{0.75\columnwidth}{!}{%
\begin{tabular}{ c | c  c  c }
net & $AP$ & $AP_{50}$ & $AP_{75}$ \\
\hline
res101-mrcn (ours)  & 30.7 & 52.3 & 32.4 \\
res101-mrcn~\cite{he2017mask} & 32.7 & 54.2 & 34.0 \\
\hline
\end{tabular}
}
\end{center}
\vspace{-15pt}
\caption{Instance segmentation results.}\label{table:segmentation}
\end{table}
 
We then compare our Mask R-CNN implementation with the original one~\cite{he2017mask} in Table~\ref{table:segmentation}. 
Note this is not a strictly fair comparison as our model was trained with fewer images.
Overall, the AP of our implementation is ${\sim}2$ points lower. 
The main reason may due to the shorter 600-pixel edge setting and smaller training batch size.
Even though, our pixel-wise comprehension results already outperform the state-of-the-art ones with a huge margin (see Table.~\ref{table:comprehension_segmentation}, and we believe there exists space for further improvements.

\subsection{More Examples}
\begin{figure*}[t]
\centering
\includegraphics[width=0.92\textwidth]{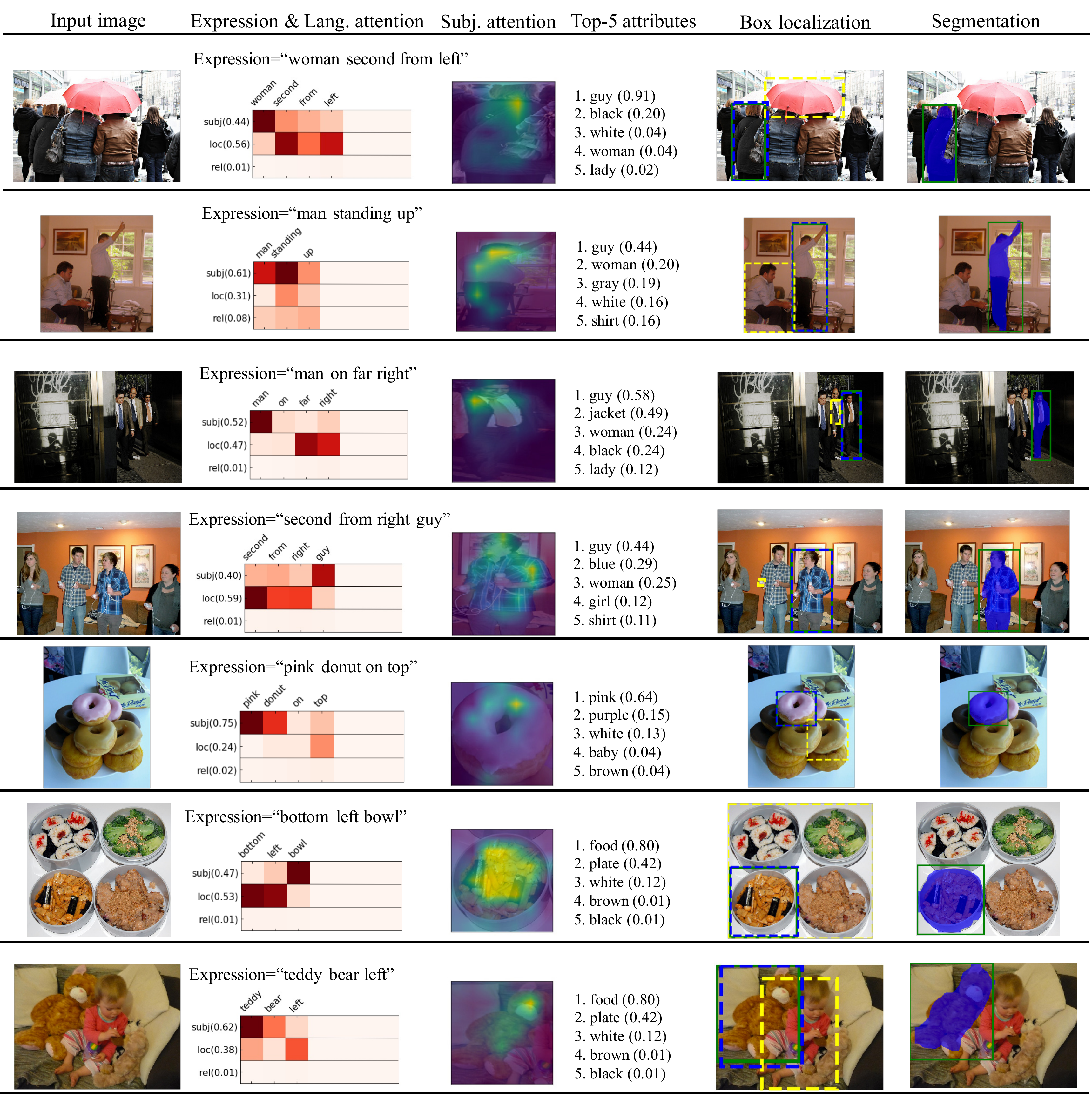}
\caption{Examples of fully automatic comprehension on \textbf{RefCOCO}. 
The 1st column shows the input image.
The 2nd column shows the expression, word attention and module weights.
The 3rd column shows our predicted subject attention, and the 4th column shows its top-5 attributes.
The 5th column shows box-level comprehension where the red dotted boxes show our prediction and yellow dotted boxes shows the relative object, and the green boxes are the ground-truth. 
The 6th column shows the segmentation.}
\label{fig:examples_correct_refcoco_unc}
\end{figure*}

\begin{figure*}[b]
\centering
\includegraphics[width=0.92\textwidth]{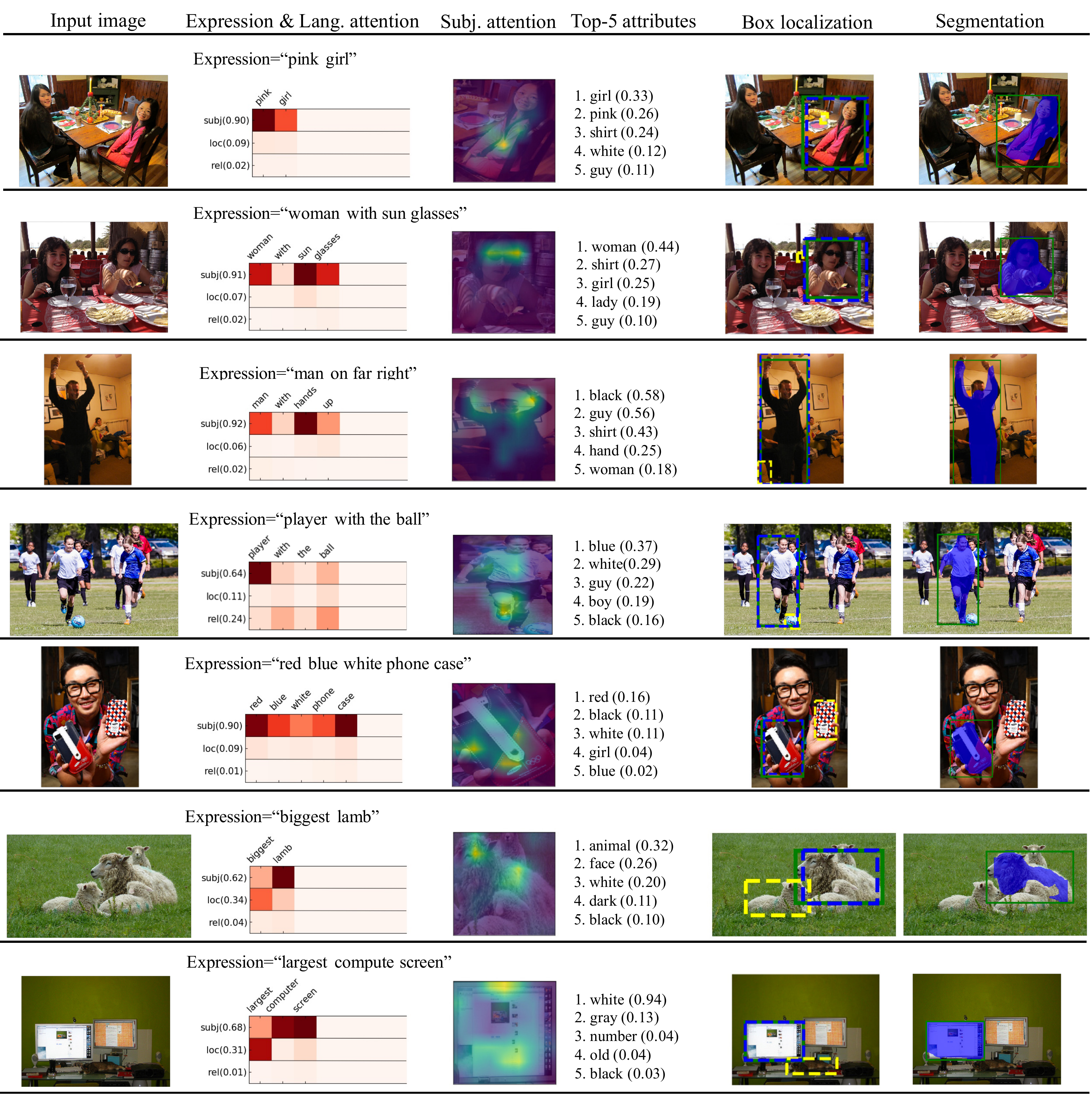}
\caption{Examples of fully automatic comprehension on \textbf{RefCOCO+}.The 1st column shows the input image.
The 2nd column shows the expression, word attention and module weights.
The 3rd column shows our predicted subject attention, and the 4th column shows its top-5 attributes.
The 5th column shows box-level comprehension where the red dotted boxes show our prediction and yellow dotted boxes shows the relative object, and the green boxes are the ground-truth. 
The 6th column shows the segmentation.}
\label{fig:examples_correct_refcoco+_unc}
\end{figure*}

\begin{figure*}[t]
\centering
\includegraphics[width=0.92\textwidth]{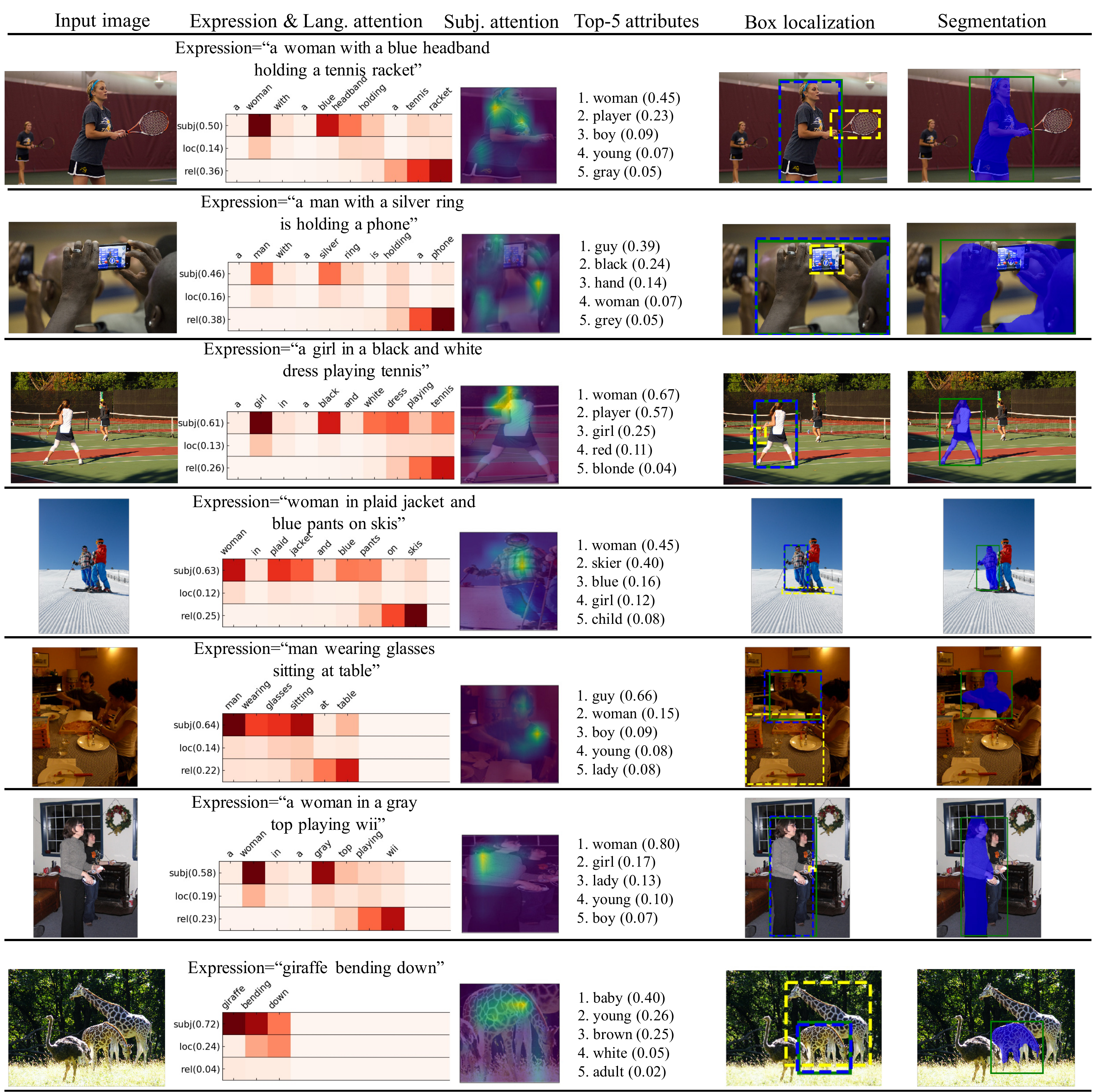}
\caption{Examples of fully automatic comprehension on \textbf{RefCOCOg}. The 1st column shows the input image.
The 2nd column shows the expression, word attention and module weights.
The 3rd column shows our predicted subject attention, and the 4th column shows its top-5 attributes.
The 5th column shows box-level comprehension where the red dotted boxes show our prediction and yellow dotted boxes shows the relative object, and the green boxes are the ground-truth. 
The 6th column shows the segmentation.}
\label{fig:examples_correct_refcocog_umd}
\end{figure*}

\begin{figure*}[t]
\centering
\includegraphics[width=0.98\textwidth]{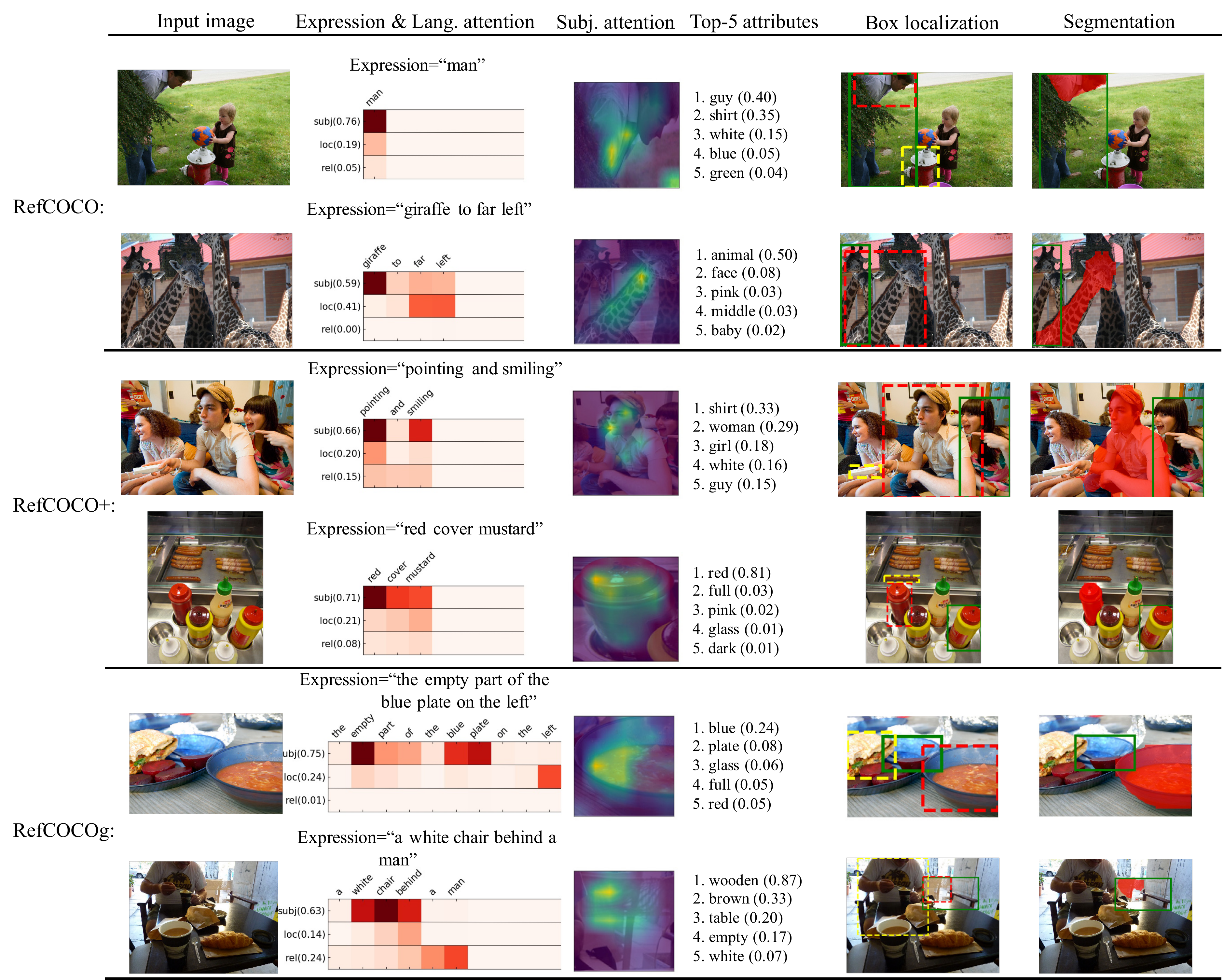}
\caption{Examples of incorrect comprehension on three datasets. The 1st column shows the input image.
The 2nd column shows the expression, word attention and module weights.
The 3rd column shows our predicted subject attention, and the 4th column shows its top-5 attributes.
The 5th column shows box-level comprehension where the red dotted boxes show our prediction and yellow dotted boxes shows the relative object, and the green boxes are the ground-truth. 
The 6th column shows the segmentation.}
\label{fig:examples_incorrect}
\end{figure*}

We show more examples of comprehension using our full model in Fig.~\ref{fig:examples_correct_refcoco_unc} (RefCOCO), Fig.~\ref{fig:examples_correct_refcoco+_unc} (RefCOCO+) and Fig.~\ref{fig:examples_correct_refcocog_umd} (RefCOCOg).
For each example, we show the input image (1st column), the input expression with our predicted module weights and word attention (2nd column), the subject attention (3rd column) and top-5 attributes (4th column), box-level comprehension (5th column), and pixel-wise segmentation (6th column).
As comparison, we also show some incorrect comprehension in Fig.~\ref{fig:examples_incorrect}.

\clearpage
\clearpage
{\small
\bibliographystyle{ieee}
\bibliography{egbib}

\begin{thebibliography}{10}\itemsep=-1pt

\bibitem{andreas2016modular}
J.~Andreas, D.~Klein, and S.~Levine.
\newblock Modular multitask reinforcement learning with policy sketches.
\newblock {\em ICML}, 2017.

\bibitem{andreas2016learning}
J.~Andreas, M.~Rohrbach, T.~Darrell, and D.~Klein.
\newblock Learning to compose neural networks for question answering.
\newblock {\em NAACL}, 2016.

\bibitem{andreas2016neural}
J.~Andreas, M.~Rohrbach, T.~Darrell, and D.~Klein.
\newblock Neural module networks.
\newblock In {\em CVPR}, 2016.

\bibitem{chen2017query}
K.~Chen, R.~Kovvuri, and R.~Nevatia.
\newblock Query-guided regression network with context policy for phrase
  grounding.
\newblock In {\em ICCV}, 2017.

\bibitem{chen2017implementation}
X.~Chen and A.~Gupta.
\newblock An implementation of faster rcnn with study for region sampling.
\newblock {\em arXiv preprint arXiv:1702.02138}, 2017.

\bibitem{he2017mask}
K.~He, G.~Gkioxari, P.~Doll{\'a}r, and R.~Girshick.
\newblock Mask r-cnn.
\newblock In {\em ICCV}, 2017.

\bibitem{he2016deep}
K.~He, X.~Zhang, S.~Ren, and J.~Sun.
\newblock Deep residual learning for image recognition.
\newblock In {\em CVPR}, 2016.

\bibitem{hu2017learning}
R.~Hu, J.~Andreas, M.~Rohrbach, T.~Darrell, and K.~Saenko.
\newblock Learning to reason: End-to-end module networks for visual question
  answering.
\newblock {\em ICCV}, 2017.

\bibitem{hu2016segmentation}
R.~Hu, M.~Rohrbach, and T.~Darrell.
\newblock Segmentation from natural language expressions.
\newblock In {\em ECCV}, 2016.

\bibitem{ronghang16relationship}
R.~Hu, M.~Rohrbacnh, J.~Andreas, T.~Darrell, and K.~Saenko.
\newblock Modeling relationship in referential expressions with compositional
  modular networks.
\newblock In {\em CVPR}, 2017.

\bibitem{hu2015natural}
R.~Hu, H.~Xu, M.~Rohrbach, J.~Feng, K.~Saenko, and T.~Darrell.
\newblock Natural language object retrieval.
\newblock {\em CVPR}, 2016.

\bibitem{johnson2017inferring}
J.~Johnson, B.~Hariharan, L.~van~der Maaten, J.~Hoffman, L.~Fei-Fei, C.~L.
  Zitnick, and R.~Girshick.
\newblock Inferring and executing programs for visual reasoning.
\newblock {\em ICCV}, 2017.

\bibitem{kazemzadeh2014referitgame}
S.~Kazemzadeh, V.~Ordonez, M.~Matten, and T.~L. Berg.
\newblock Referitgame: Referring to objects in photographs of natural scenes.
\newblock In {\em EMNLP}, 2014.

\bibitem{lin2014microsoft}
T.-Y. Lin, M.~Maire, S.~Belongie, J.~Hays, P.~Perona, D.~Ramanan,
  P.~Doll{\'a}r, and C.~L. Zitnick.
\newblock Microsoft coco: Common objects in context.
\newblock In {\em ECCV}, 2014.

\bibitem{liu2017recurrent}
C.~Liu, Z.~Lin, X.~Shen, J.~Yang, X.~Lu, and A.~Yuille.
\newblock Recurrent multimodal interaction for referring image segmentation.
\newblock In {\em ICCV}, 2017.

\bibitem{liu2017referring}
J.~Liu, L.~Wang, and M.-H. Yang.
\newblock Referring expression generation and comprehension via attributes.
\newblock In {\em ICCV}, 2017.

\bibitem{long2015fully}
J.~Long, E.~Shelhamer, and T.~Darrell.
\newblock Fully convolutional networks for semantic segmentation.
\newblock In {\em CVPR}, 2015.

\bibitem{luo2017comprehension}
R.~Luo and G.~Shakhnarovich.
\newblock Comprehension-guided referring expressions.
\newblock {\em CVPR}, 2017.

\bibitem{mao2015generation}
J.~Mao, J.~Huang, A.~Toshev, O.~Camburu, A.~Yuille, and K.~Murphy.
\newblock Generation and comprehension of unambiguous object descriptions.
\newblock {\em CVPR}, 2016.

\bibitem{nagaraja16refexp}
V.~K. Nagaraja, V.~I. Morariu, and L.~S. Davis.
\newblock Modeling context between objects for referring expression
  understanding.
\newblock In {\em ECCV}, 2016.

\bibitem{ren2015faster}
S.~Ren, K.~He, R.~Girshick, and J.~Sun.
\newblock Faster r-cnn: Towards real-time object detection with region proposal
  networks.
\newblock In {\em NIPS}, 2015.

\bibitem{rohrbach2016grounding}
A.~Rohrbach, M.~Rohrbach, R.~Hu, T.~Darrell, and B.~Schiele.
\newblock Grounding of textual phrases in images by reconstruction.
\newblock In {\em ECCV}, 2016.

\bibitem{simonyan2014very}
K.~Simonyan and A.~Zisserman.
\newblock Very deep convolutional networks for large-scale image recognition.
\newblock {\em arXiv preprint arXiv:1409.1556}, 2014.

\bibitem{socher2013parsing}
R.~Socher, J.~Bauer, C.~D. Manning, et~al.
\newblock Parsing with compositional vector grammars.
\newblock In {\em ACL}, 2013.

\bibitem{su2017reasoning}
J.-C. Su, C.~Wu, H.~Jiang, and S.~Maji.
\newblock Reasoning about fine-grained attribute phrases using reference games.
\newblock {\em ICCV}, 2017.

\bibitem{wang2015learning}
L.~Wang, Y.~Li, and S.~Lazebnik.
\newblock Learning deep structure-preserving image-text embeddings.
\newblock {\em CVPR}, 2016.

\bibitem{wang2017learning}
L.~Wang, Y.~Li, and S.~Lazebnik.
\newblock Learning two-branch neural networks for image-text matching tasks.
\newblock {\em arXiv preprint arXiv:1704.03470}, 2017.

\bibitem{wu2017image}
Q.~Wu, C.~Shen, P.~Wang, A.~Dick, and A.~van~den Hengel.
\newblock Image captioning and visual question answering based on attributes
  and external knowledge.
\newblock {\em IEEE Transactions on Pattern Analysis and Machine Intelligence},
  2017.

\bibitem{yang2016hierarchical}
Z.~Yang, D.~Yang, C.~Dyer, X.~He, A.~J. Smola, and E.~H. Hovy.
\newblock Hierarchical attention networks for document classification.
\newblock In {\em HLT-NAACL}, 2016.

\bibitem{yao2016boosting}
T.~Yao, Y.~Pan, Y.~Li, Z.~Qiu, and T.~Mei.
\newblock Boosting image captioning with attributes.
\newblock {\em arXiv preprint arXiv:1611.01646}, 2016.

\bibitem{you2016image}
Q.~You, H.~Jin, Z.~Wang, C.~Fang, and J.~Luo.
\newblock Image captioning with semantic attention.
\newblock In {\em CVPR}, 2016.

\bibitem{yu2016refer}
L.~Yu, P.~Poirson, S.~Yang, A.~C. Berg, and T.~L. Berg.
\newblock Modeling context in referring expressions.
\newblock In {\em ECCV}, 2016.

\bibitem{yu2016joint}
L.~Yu, H.~Tan, M.~Bansal, and T.~L. Berg.
\newblock A joint speaker-listener-reinforcer model for referring expressions.
\newblock In {\em CVPR}, 2017.

\end{thebibliography}
}

\end{document}